\documentclass[10pt,twocolumn,letterpaper]{article}

\usepackage{cvpr}
\usepackage{times}
\usepackage{epsfig}
\usepackage{graphicx}
\usepackage{amsmath}
\usepackage{amssymb}
\usepackage{subfigure}
\usepackage{multirow}
\usepackage{siunitx}

\usepackage[pagebackref=true,breaklinks=true,letterpaper=true,colorlinks,bookmarks=false]{hyperref}

\cvprfinalcopy 

\def\cvprPaperID{2920} 

\ifcvprfinal\fi
\begin{document}

\title{Partial Convolution based Padding}

\author{Guilin Liu \quad Kevin J. Shih \quad Ting-Chun Wang \quad Fitsum A. Reda\\
Karan Sapra \quad Zhiding Yu \quad Andrew Tao \quad Bryan Catanzaro\\
NVIDIA\\
{\tt\small \{guilinl, kshih, tingchunw, freda, ksapra, zhidingy, atao, bcatanzaro\}@nvidia.com}}


\maketitle

\begin{abstract}
In this paper, we present a simple yet effective padding scheme that can be used as a drop-in module for existing convolutional neural networks. We call it \textbf{partial convolution based padding}, with the intuition that the padded region can be treated as holes and the original input as non-holes. Specifically, during the convolution operation, the convolution results are re-weighted near image borders based on the ratios between the padded area and the convolution sliding window area. Extensive experiments with various deep network models on ImageNet classification and semantic segmentation demonstrate that the proposed padding scheme consistently outperforms standard zero padding with better accuracy. The code is available at \href{https://github.com/NVIDIA/partialconv}{https://github.com/NVIDIA/partialconv}
\end{abstract}

\section{Introduction}

Convolutional operation often requires padding when part of the filter extends beyond the input image or feature map. Standard approaches include zero padding (extend with zeros), reflection padding (reflect the input values across the border axis) and replication padding (extend by replicating the values along borders).
Among them, the most commonly used scheme is zero padding, as was adopted by~\cite{krizhevsky2012imagenet}. Besides its simplicity, zero padding is also computationally more efficient than the other two schemes. Yet, there is no consensus on which padding scheme is the best yet. In this work, we conduct extensive experiments on the ImageNet classification task using these three padding schemes, and found that reflection padding and replication padding can get similar or sometimes worse performance compared with zero padding. While these three padding schemes allow the convolution to safely operate along the borders of the input, they incorporate extrapolated information, which may adversely affect the model's quality. 

\begin{figure}
    \centering
    \includegraphics[width=0.5\textwidth]{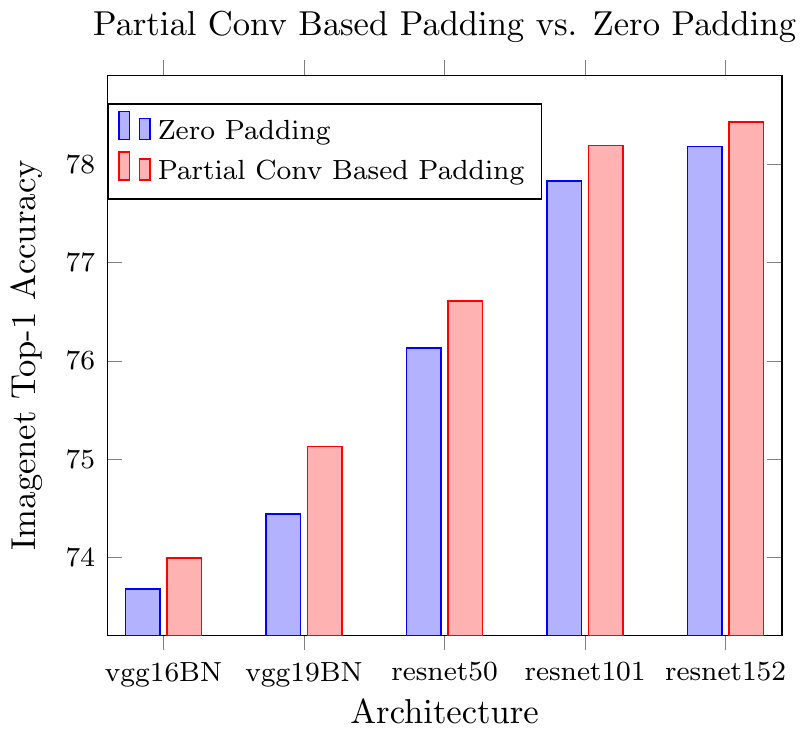}
    \caption{Comparison of the ImageNet classification top-1 accuracy with center crop between partial convolution based padding (in red) and zero padding (in blue) on VGG and ResNet networks. vgg16BN and vgg19BN represent the vgg16 network and vgg19 network with batch normalization layers.}
    \label{fig:teaser}
\end{figure}

Each of the three existing standard padding approaches makes assumptions that may have undesirable effects to the model quality in different ways. Zero padding works like adding extra unrelated data to the input. Reflection and replication padding attempt to pad with plausible data values by re-using what is along the borders of the input. These two types of padding lead to unrealistic image patterns since only some parts of the input are replicated. Moreover, for all these three padding schemes, the added or replicated features are treated equally as the original input, which makes it possible for the network to be confused.

To eliminate the potential undesired effects from the extrapolated inputs, we propose a new padding scheme called \textit{partial convolution based padding}, which conditions the convolution output only on the valid inputs. In contrast to the zero padding where zeros are used for the missing inputs, our partial convolution based padding adaptively re-weights the output to adjust for the fraction of the missing data. To this end, our padding scheme treats the padded region as holes and applies the partial convolution layer~\cite{liu2018image} for the re-weighting in the convolution operation. We demonstrate through experiments on ImageNet classification that VGG and ResNet architectures gain better accuracy using partial convolution based padding than using zero padding.
Furthermore, we observed that models trained with zero padding are very sensitive to the padding used during inference, and would have a big performance drop if different padding is used. This suggests that when zero padding is used, part of the parameters in the model are wasted dealing with the padded data. On the other hand, models trained with partial convolution based padding are robust, and perform similarly no matter which padding is used during inference.

Our main contributions can be summarized as follows: 
\begin{enumerate}
\item We conduct extensive experiments to show that by replacing the zero padding with partial convolution based padding on various models and tasks, we obtain better results and faster convergence with fewer training iterations.
\item We show that among the three existing padding schemes, reflection padding and replication padding perform similarly with or worse than zero padding on ImageNet classification task. Padding scheme like zero padding is sensitive to the particular padding used during training. On the other hand, our partial convolution based padding are robust to the input padding type.
\item We demonstrate that partial convolution based padding can improve semantic segmentation on regions near image boundaries, suggesting that existing padding techniques result in lower accuracy as more of the receptive field is conditioned on padded values.
\end{enumerate}

\section{Related Work}
Researchers have tried to improve the performance of CNN models from almost all the aspects including different variants of SGD optimizer (SGD, Adam~\cite{kingma2014adam}, RMSprop~\cite{tieleman2012lecture}, Adadelta~\cite{zeiler2012adadelta}), non-linearity layers (ReLU~\cite{nair2010rectified}, LeakyReLU~\cite{maas2013rectifier}, PReLU~\cite{he2015delving}), normalization layers (Batch Norm~\cite{ioffe2015batch}, Instance Norm~\cite{ulyanov1607instance}, Layer Norm~\cite{ba2016layer}, Group Norm~\cite{wu2018group}), etc. However, little attention has been paid to improving the padding schemes. Innamorati et al.~\cite{innamorati2018learning} augments networks with a separate set of filters to explicitly handle boundaries. Specifically, it learns a total of $9$ filters for a convolutional layer, one for the middle part and the rest for boundary cases. Cheng et al.~\cite{cheng2018cube} propose a special image projection to handle undefined boundaries resulting from projecting a \ang{360} view image onto a 2D image. Images are first projected onto a cube and a final image is formed by concatenating cube faces. As such, large undefined borders can be eliminated. Our proposed padding scheme is orthogonal to all of these tricks and can be used in conjunction with them. It also does not increase the parameter count; instead, it uses a single filter with a simple yet effective re-weighted convolution at boundary pixels. 

\textbf{Deep learning for inputs with missing data}. In our setup, the padded regions are interpreted as missing data. Handling inputs with missing data is related to the problems of image inpainting and other data completion tasks. Recent deep learning based methods \cite{pathak2016context,li2017generative,iizuka2017globally,yang2017high} initialize the holes (regions with missing data) with some constant placeholder values and treat the newly initialized hole region and original non-hole region equally. Then GAN based networks will be used to re-generate a new image conditioned on this initialized image. The results usually suffer from dependence on the initial hole values, with lack of texture in hole regions, obvious color contrasts or artificial edge responses. Many of them require post-processing, like a refinement network~\cite{yu2018generative,song2017image}, Poisson blending~\cite{iizuka2017globally} or other post-optimization techniques~\cite{yang2017high}. Some other methods~\cite{ulyanov2017deep,yeh2016semantic,uhrig2017sparsity} ignore the hole placeholder values in the image inpainting or data completion tasks; none of them has explicitly handled the missing data at the padded regions yet. 

\textbf{Reweighted convolution}. Reweighted convolution as the variant of typical convolution has been explored in several tasks. Harley et al.~\cite{harley2017segmentation} uses soft attention masks to reweight the convolution results for semantic segmentation. PixelCNN designs the reweighted convolution such that the next pixel generation only depends on previous generated pixels. Inpainting methods like~\cite{uhrig2017sparsity,ren2015shepard} take the hole mask into consideration for reweighting the convolution results. For all these methods, none of their corresponding reweighting mechanisms has been used to handle the padding yet.

\section{Formulation \& Analysis}

In this section, we start with reviewing the partial convolution~\cite{liu2018image} and then illustrate the idea of partial convolution based padding. 

\begin{figure*}[h]
\centering
 \scalebox{0.8}{
{
    \centering
    \subfigure[$\mathbf{X}$]{\label{fig:X}{\includegraphics[width=0.2\textwidth]{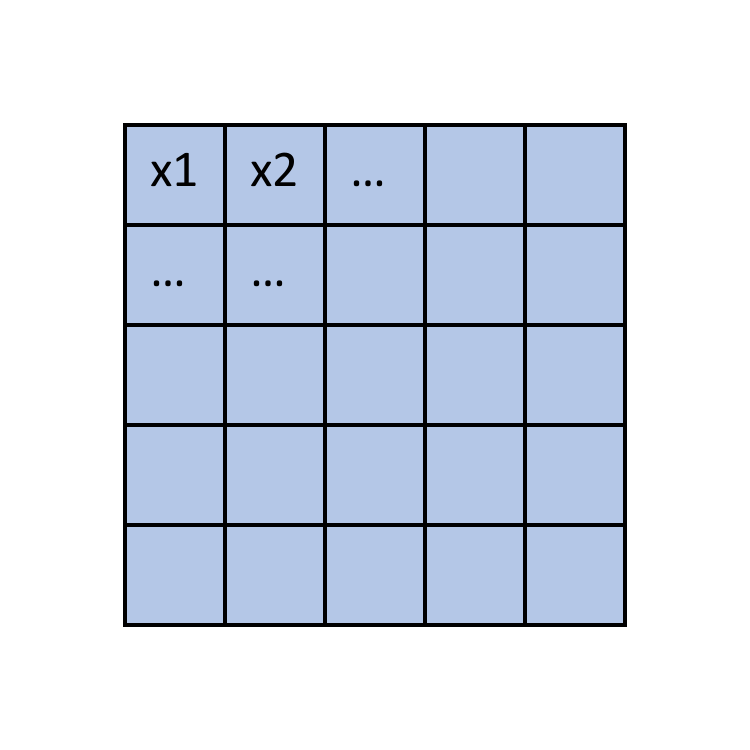}}}
    \subfigure[$\mathbf{1}$]{\label{fig:1}{\includegraphics[width=0.2\textwidth]{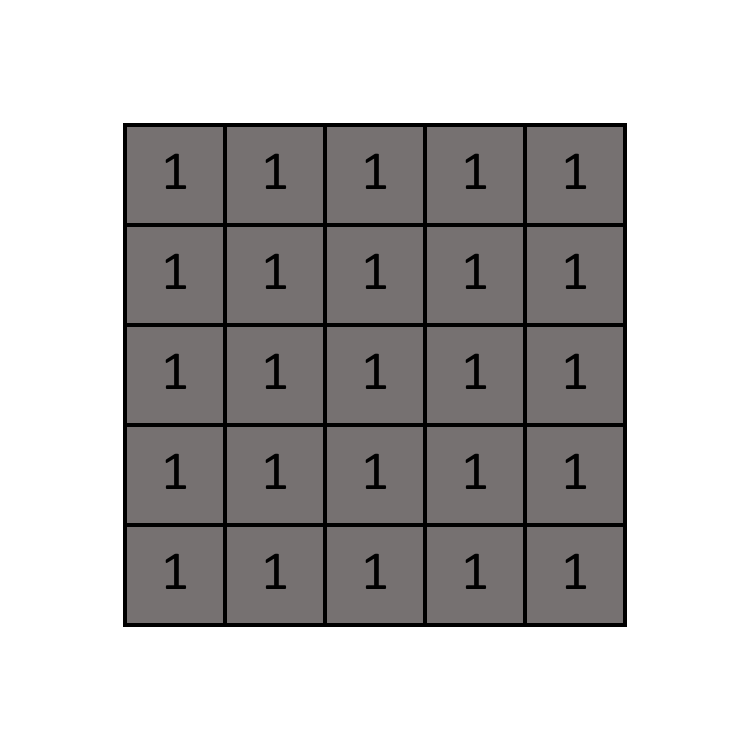}}}
    \subfigure[$\mathbf{X}^{p0}$]{\label{fig:x_p0}{\includegraphics[width=0.2\textwidth]{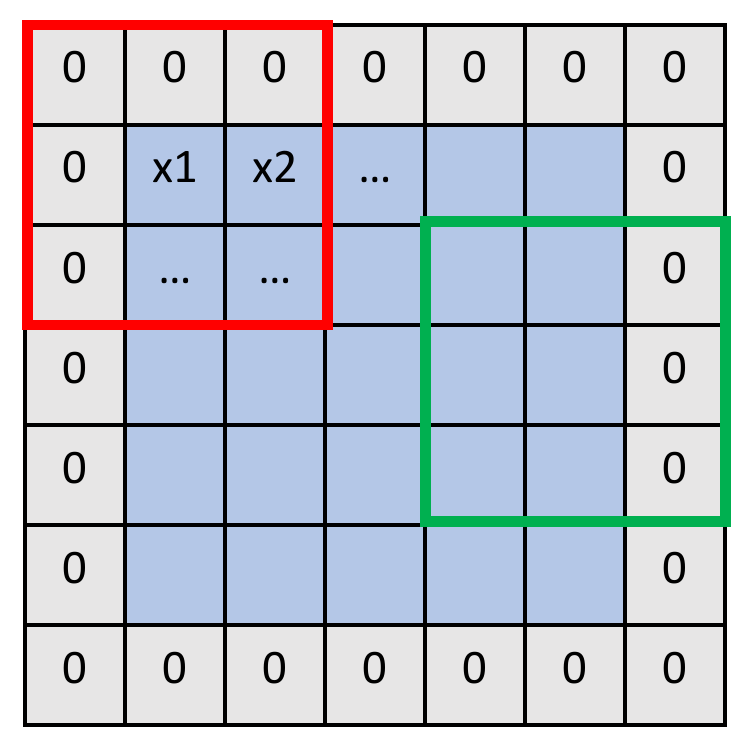}}}
    \subfigure[$\mathbf{1}^{p0}$]{\label{fig:1_p0}{\includegraphics[width=0.2\textwidth]{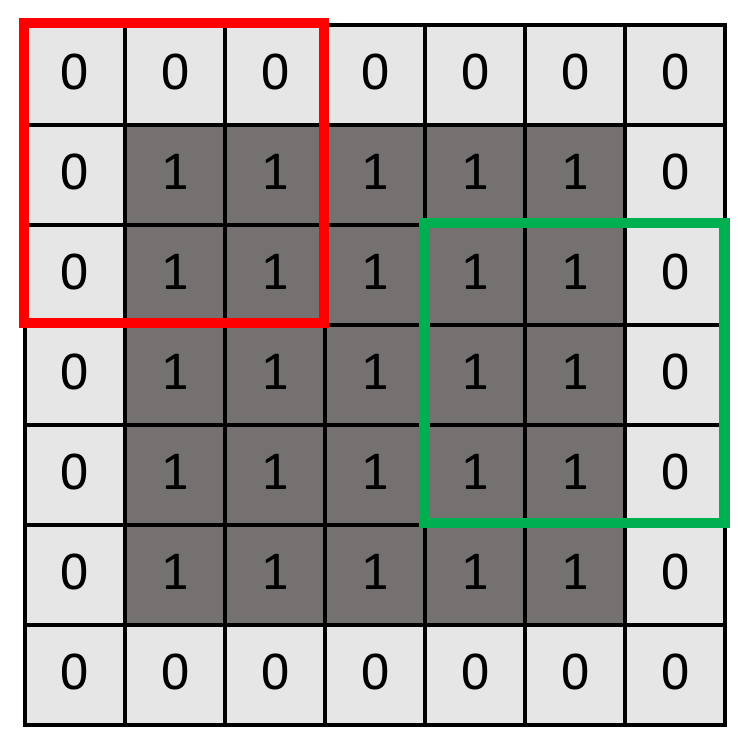}}}
    \subfigure[$\mathbf{1}^{p1}$]{\label{fig:1_p1}{\includegraphics[width=0.2\textwidth]{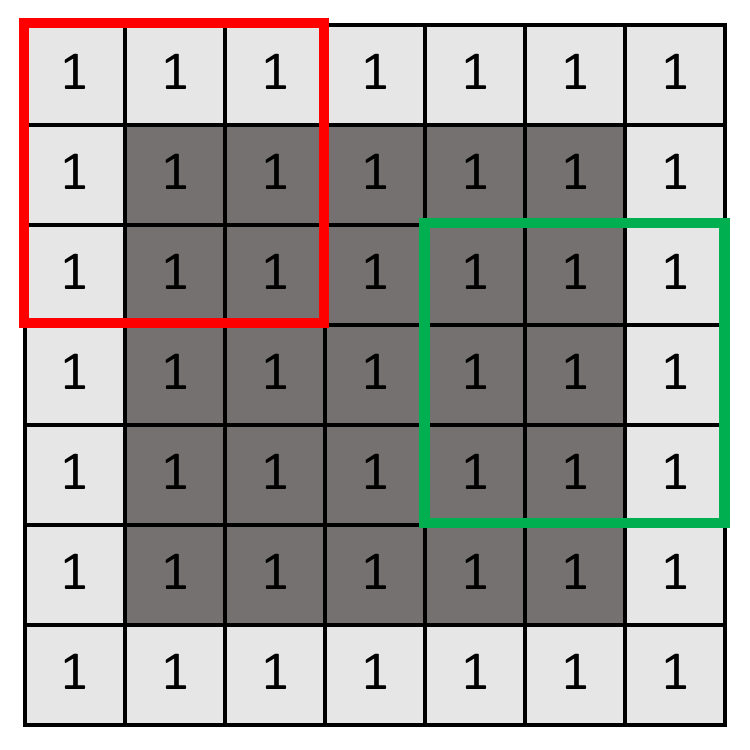}}}
}
}
    \caption{Visualization of $\mathbf{X}$, $\mathbf{1}$, $\mathbf{X}^{p0}$, $\mathbf{1}^{p0}$ and $\mathbf{1}^{p1}$; the red and green boxes are the sliding convolution window examples centering at posiiton $(i, j)$. The result of convolution with typical zero padding will only depend on $\mathbf{X}^{p0}_{(i,j)}$: $\mathbf{W}^T\mathbf{X}^{p0}_{(i,j)}+b$ ; the result with partial convolution based padding will rely on both $\mathbf{X}^{p0}_{(i,j)}$ and $\mathbf{1}^{p0}_{(i,j)}$: $\mathbf{W}^T\mathbf{X}^{p0}_{(i,j)}\frac{||\mathbf{1}^{p1}_{(i,j)}||_1}{||\mathbf{1}^{p0}_{(i,j)}||_1} + b$.}
    \label{fig:notation}
    \vspace{-.2cm}
\end{figure*}

\textbf{Partial Convolution}. Partial convolution~\cite{liu2018image} is originally proposed to handle incomplete input data, such as images with holes. Let $\mathbf{X}_{(i,j)}$ be the feature values (pixel values) for the current convolution (sliding) window at the position $(i,j)$ and $\mathbf{M}_{(i,j)}$ be the corresponding binary mask with the hole region being 0 and non-hole region being 1. The partial convolution (ignoring bias) 
at every location is defined as:

\begin{equation}
\label{eq:partconv}
    x'_{(i,j)} = \begin{cases}
     \mathbf{W}^T(\mathbf{X}_{(i,j)}\odot\mathbf{M}_{(i,j)})r_{(i,j)},&||\mathbf{M}_{(i,j)}||_{1}>0  \\
        0,&\text{otherwise}
    \end{cases}
\end{equation}

where
\begin{equation}
    r_{(i,j)} = \frac{||\mathbf{1}_{(i,j)}||_1}{||\mathbf{M}_{(i,j)}||_1},
\end{equation}
$\odot$ denotes element-wise multiplication, $\mathbf{1}_{(i,j)}$ is the all-one vector with the same shape as $\mathbf{M}_{(i,j)}$ and $\mathbf{W}$ is the filter weight matrix. We compute $x'_{(i,j)}=x'_{(i,j)}+b$ to account for an additional bias term (when $||\mathbf{M}_{(i,j)}||_{1}>0$). As can be seen, output values depend only on the unmasked inputs. The scaling factor $||\mathbf{1}_{(i,j)}||_1/||(\mathbf{M}_{(i,j)}||_1$ applies appropriate scaling to adjust for the varying amount of valid (unmasked) inputs.

After each partial convolution operation, we then update our mask as follows: if the convolution was able to condition its output on at least one valid input value, then we mark that location to be valid. This is expressed as:
\begin{equation}
m'_{(i,j)}=\begin{cases}
    1, & \text{if } ||\mathbf{M}_{(i,j)}||_{1}>0  \\
    0, & \text{otherwise}
\end{cases}
\label{eq:mask_update}
\end{equation}
and can easily be implemented in any deep learning framework as part of the forward pass. With sufficient successive applications of the partial convolution layer, any mask will eventually be all ones, if the input contained any valid pixels.

\textbf{Partial Convolution based Padding}.
Partial convolution is extended to handle padding by defining the input region to be non-hole and padded region to be holes. Specifically,
given the input image/feature $\mathbf{X}_{(i,j)}$ at the border, let $\mathbf{1}_{(i,j)}$ denotes the 2D matrix having same height and width as $\mathbf{X}_{(i,j)}$, but with a single channel; $\mathbf{X}^{p0}_{(i,j)}$ denotes the zero padded result of $\mathbf{X}_{(i,j)}$; $\mathbf{1}^{p0}_{(i,j)}$ denotes the zero padded result of $\mathbf{1}_{(i,j)}$; $\mathbf{1}^{p1}_{(i,j)}$ denotes the one padded result of $\mathbf{1}_{(i,j)}$; the visualization of their examples can be found in Figure~\ref{fig:notation}; then the convolution result $x'_{(i,j)}$ is computed as following:

\begin{equation}
\label{eq:partconv}
    x'_{(i,j)} = \mathbf{W}^T(\mathbf{X}^{p0}_{(i,j)}\odot\mathbf{1}^{p0}_{(i,j)})r_{(i,j)} + b = \\ \mathbf{W}^T\mathbf{X}^{p0}_{(i,j)}r_{(i,j)} + b
\end{equation}
where
\begin{equation}
    r_{(i,j)} = \frac{||\mathbf{1}^{p1}_{(i,j)}||_1}{||\mathbf{1}^{p0}_{(i,j)}||_1}
\end{equation}

Based on Equation~\ref{eq:partconv}, we can see that the widely used zero padding is a special case of our partial convolution based padding by directly setting $r_{(i,j)}$ to be 1, which can be formulated as following:

\begin{equation}
\label{eq:convzeropadding}
x'=\mathbf{W}^T\mathbf{X}^{p0}_{(i,j)}+b
\end{equation}

\textbf{Case of Big Padding Size}. 
 In some cases, big padding will be needed. For example, input images may have different image sizes; big paddings may thus be needed to make images uniformly sized before feeding them into the network. Re-sizing by padding is typically preferred because normal re-sizing may introduce distortions due to altered aspect ratio. Convolution at such borders may not have valid/original data because the window size could be smaller than the padding size. 
For such cases, we follow the original partial convolution formulation in~\cite{liu2018image} to include the mask updating step. Specifically, for the very first convolutional layer, the input mask $\mathbf{M}_{1st-layer}$ will be the padded result $\mathbf{1}^{p0}$ ($\mathbf{M}_{1st-layer}$ = $\mathbf{1}^{p0}$); and it will produce an output mask $\mathbf{M}'_{1st-layer}$ using the rule in Equation~\ref{eq:mask_update}. The mask input for the next layer will be the padded result of $\mathbf{M}'_{1st-layer}$, namely $\mathbf{M}_{2nd-layer}$ = $\mathbf{M}'^{\ p0}_{1st-layer}$. Thus, the input mask for $n-th$ layer will be $\mathbf{M}_{nth-layer}$ = $\mathbf{M}'^{\ p0}_{(n-1)th-layer}$.

\section{Implementation \& Experiments}

\textbf{Implementation Details}. We provide a pure-PyTorch implementation of the convolution layer with the proposed padding scheme on top of existing PyTorch modules. We implement the mask of ones ($\mathbf{1}$) as a single-channel feature with the same batch size, height and width as the input tensor $\mathbf{X}$. The 1-padded version of $\mathbf{1}$ ($||\mathbf{1}_{p1}||_1$) is directly set to be $k_h * k_w$, where $k_h$ and $k_w$ are the height and width of the kernel. $||\mathbf{1}_{p0}||_1$ is implemented by calling the convolution operation once with all the weights being 1, bias being 0 and original target padding size. The result of $\frac{||\mathbf{1}_{p1}||_1}{||\mathbf{1}_{p0}||_1}$ only needs to be computed at the first time and the result can be cached for future iterations as long as  the input size does not change. Thus, the execution time starting from the second iteration will be lower than the first iteration.  In Table~\ref{tab:running_time}, We show the comparison of GPU execution time between zero padding powered networks and partial convolution based padding powered networks for both the first iteration and after iterations. It can be seen that starting from the second iteration partial convolution powered vgg16 (batch normalization layer version) and resnet50 only cost about $4$\% more time to do the inference on a $224 \times 224$ input image with a single NVIDIA V100 GPU. Note that this current implementation is based on existing PyTorch interfaces to allow users to use without compiling new CUDA kernel. If this padding scheme is implemented using CUDA directly, the extra cost would be negligible as we only need to re-weight the conovolution results at the border. 
\begin{table}[h]
    \centering
    \begin{tabular}{l|c|c}
          & running time & relative  \\
        \hline
        vgg16BN\_zero & 13.791 & 100\% \\
        \hline
        vgg16BN\_partial ($1$st) & 14.679 & 106.44\% \\
        vgg16BN\_partial ($2$nd - $n$th)  & 14.351 & 104.06\% \\
        \hline
        \hline
        resnet50\_zero & 5.779 & 100\% \\
        \hline
        resnet50\_partial ($1$st) & 7.496 & 129.71\% \\
        resnet50\_partial ($2$nd - $n$th) & 5.975 & 103.39\% \\
        \hline
    \end{tabular}
    \caption{The inference time (measured in ms) comparison between zero padding powered networks and partial convolution based padding powered networks ( VGG16 network with Batch Normalization layers and ResNet50) using a 224x224 image as input. vgg16BN and resnet50 are with the default zero padding while vgg16BN\_partial and resnet50\_partial are with partial convolution based padding.This is based on the raw PyTorch implementation without implementing custom CUDA kernels. The time is measured on a single NVIDIA V100 GPU. $1$st, $2$nd and $n$th represent the first iteration, second iteration and $n$th iteration respectively.}
    \label{tab:running_time}
\end{table}

\begin{table*}
    \centering
    \begin{tabular}{c|ccccc|ccc||c}
Network&1\_best&2\_best&3\_best&4\_best&5\_best&average&diff&stdev & PT\_official \\
\hline
vgg16BN\_zero&73.986&73.826&73.610&73.778&73.204&73.681& - &0.244 & 73.37 \\
vgg16BN\_partial&74.154&74.072&73.790&73.898&74.072&\textbf{73.997}&0.316&0.121 & - \\
vgg16BN\_ref&73.404&73.670&73.364& 73.428 & 73.416 &73.456&-0.225&0.122& -\\
vgg16BN\_rep&73.638&73.580&73.816&74.012 &73.734&73.756 &0.075&0.169& - \\
\hline
vgg19BN\_zero&74.378&74.838&74.168&74.236&74.588&74.442& - &0.224 & 74.24 \\
vgg19BN\_partial&74.940&75.288&75.092&75.160&75.164&\textbf{75.129}&0.687&0.104 & - \\
vgg19BN\_ref&74.260 & 74.118 & 73.330 & 74.042 & 74.268 & 74.004 &-0.438&0.389& - \\
vgg19BN\_rep&74.706 & 74.120 & 74.352 &74.182& 74.290 &74.330&-0.112&0.229& - \\
\hline
resnet50\_zero&76.240&76.074&75.988&76.136&76.224&76.132& - &0.086 & 76.15 \\
resnet50\_partial&76.606&76.532&76.638&76.562&76.716&\textbf{76.611}&0.478&0.058 & - \\
resnet50\_ref & 76.188 & 76.24 & 76.38 & 75.884 & 76.182 & 76.174 & 0.042 & 0.181 & - \\
resnet50\_rep & 76.048 & 76.452 & 76.19 & 76.186 & 76.158 & 76.207 & 0.075 & 0.149 & - \\
\hline
resnet101\_zero&77.942&77.880&77.428&77.868&78.044&77.832& - &0.193 & 77.37 \\
resnet101\_partial&78.318&78.124&78.166&78.090&78.264&\textbf{78.192}&0.360&0.078 & - \\
resnet101\_ref&77.896&77.75&78.026&77.584&78.004&77.852&0.02&0.185&- \\
resnet101\_rep&78.022&77.706&77.928&78.1&77.758&77.903&0.071&0.168& - \\
\hline
resnet152\_zero&78.364&78.164&78.130&78.018&78.242&78.184& - &0.105 & 78.31 \\
resnet152\_partial&78.514&78.338&78.252&78.516&78.540&\textbf{78.432}&0.248&0.105 & - \\
resnet152\_ref & 77.472 & 78.046 & 77.862 & 77.682 & 77.962 & 77.805 & -0.379 & 0.230 & - \\
resnet152\_rep & 77.852 & 78.308 & 78.088 & 77.924 & 77.948 & 78.024 & -0.16 & 0.180 & - \\
    \end{tabular}
    \caption{The best top-1 accuracies for each run with 1-crop testing. *\_zero, *\_partial, *\_ref and *\_rep indicate the corresponding model with zero padding, partial convolution based padding, reflection padding and replication padding respectively. *\_best means the best validation score for each run of the training. Column average represents the average accuracy of the 5 runs. Column diff represents the difference with corresponding network using zero padding. Column stdev represents the standard deviation of the accuracies from 5 runs. PT\_official represents the corresponding official accuracies published on PyTorch website: https://pytorch.org/docs/stable/torchvision/models.html}
    \label{tab:best}
\end{table*}

\begin{table*}
    \centering
    \begin{tabular}{c|ccccc|ccc}
Network&1\_last5$_{avg}$&2\_last5$_{avg}$&3\_last5$_{avg}$&4\_last5$_{avg}$&5\_last5$_{avg}$&average&diff&stdev \\
\hline
vgg16BN\_zero&73.946&73.771&73.488&73.724&73.166&73.619& - &0.246 \\
vgg16BN\_partial&74.135&74.032&73.718&73.860&73.990&\textbf{73.947}&0.328&0.132 \\
vgg16BN\_ref&73.375&73.534 & 73.303 & 73.308 & 73.367 & 73.377 & -0.242 & 0.093 \\
vgg16BN\_rep&73.581 & 73.540 & 73.767 & 73.948 & 73.694 & 73.706 & 0.087 & 0.162 \\
\hline
vgg19BN\_zero&74.349&74.774&74.090&74.189&74.540&74.388& - &0.224 \\
vgg19BN\_partial&74.867&75.196&75.037&75.116&75.141&\textbf{75.071}&0.683&0.104 \\
vgg19BN\_ref& 74.215 & 73.980 & 73.281 & 73.954 & 74.183 & 73.923 & -0.415 & 0.377 \\
vgg19BN\_rep& 74.665 & 74.079 & 74.241 & 74.121 & 74.185 & 74.258 & -0.130 & 0.236 \\
\hline
resnet50\_zero&76.178&76.000&75.869&76.048&76.161&76.051& - &0.103 \\
resnet50\_partial&76.554&76.495&76.553&76.550&76.677&\textbf{76.566}&0.514&0.055 \\
resnet50\_ref&76.067 & 76.157 & 76.342 & 75.818 & 76.132 & 76.103 & 0.052 & 0.189 \\
resnet50\_rep&75.991 & 76.376 & 76.109 & 76.145 & 76.081 & 76.140 & 0.089 & 0.143 \\
\hline
resnet101\_zero&77.899&77.810&77.235&77.820&77.996&77.752& - &0.244 \\
resnet101\_partial&78.239&78.032&78.144&78.066&78.198&\textbf{78.136}&0.384&0.071 \\
resnet101\_ref&77.840 & 77.6812 & 77.989 & 77.458 & 77.923 & 77.778 & 0.026 & 0.213\\
resnet101\_rep&77.964 & 77.626 & 77.858 & 78.037 & 77.719 & 77.841 & 0.089 & 0.169 \\
\hline
resnet152\_zero&78.302&78.038&78.011&77.944&77.926&78.044& - &0.124 \\
resnet152\_partial&78.396&78.175&78.168&78.455&78.414&\textbf{78.322}&0.277&0.113 \\
resnet152\_ref & 77.36 & 78.01 & 77.756 & 77.616 & 77.848 & 77.718 & -0.326 & 0.246 \\
resnet152\_rep & 77.693 & 78.197 & 78.020 & 77.58 & 77.865 & 77.871 & -0.173 & 0.247 \\
    \end{tabular}
    \caption{The mean of last 5 epochs' top-1 accuracy ($\%$) for each run with 1-crop testing. *\_zero, *\_partial, *\_ref and *\_rep indicate the corresponding model with zero padding, partial convolution based padding, reflection padding and replication padding respectively. *\_last5 means the mean validation score of the last 5 epochs' model checkpoints for each run of the training. Column average represents the average accuracy of the 5 runs. Column diff represents the difference with corresponding network using zero padding. Column stdev represents the standard deviation of the accuracies from 5 runs.}
    \label{tab:avg}
\end{table*}

\subsection{Image Classification Networks}

\textbf{Experiments}. We conduct the experiments and comparisons on the ImageNet classification tasks. We train the \textit{VGG16, VGG19}~\cite{simonyan2014very}, \textit{ResNet50, ResNet101, ResNet152}~\cite{he2016deep} models on ImageNet training dataset and evaluate on the ImageNet validation dataset. We use the versions of VGG16 and VGG19 with Batch Normalization layers, as the VGG network with batch normalization is known to be less sensitive to learning rates and initializations, thereby reducing the amount of variation in our experiments. We use the raining scripts and model definitions from the corresponding official PyTorch examples. For each model, we replace all the zero padding with partial convolution based padding while keeping all the other settings and training parameters the same. While the default PyTorch training script trains for $90$ epochs, we use $100$ in our experiments like~\cite{wu2018group}. The models are all trained with $8$ NVIDIA V100 GPUs on DGX-1 workstations. The initial learning rate is set to be $0.1$ and decreased by a factor of $10$ at $30$, $60$ and $90$ epochs. We train each network $5$ times to account for variances due to initializations and stochastic mini batch sampling. To have a full comparison with all the existing padding schemes, we also run the experiments for reflection padding and replication padding with the same setting. This gives us a total of $100$ sets of trained ImageNet model weights ($20$ architectures $\times$ $5$). For each training run, we select the checkpoint with the highest validation set accuracy to represent that run (referred to as ``best"). 

In addition, we also report results using the experimental procedures from Wu et al.~\cite{wu2018group}. Specifically, Wu et al.~\cite{wu2018group} trained the network once and evaluates the performance on the model and accounts for model variance by reporting the average performance of the final $5$ epochs. As such, we also report results in which we average across the last $5$ epochs for each run (referred to as ``last\_$5_{avg}$").

\textbf{Analysis}. We evaluate the models using the top-1 classification accuracy based on the single center crop evaluation. Table~\ref{tab:best} shows the results for the ``best" scenario, whereas Table~\ref{tab:avg} shows the results for the ``last\_$5_{avg}$" setting. Note that the top-1 accuracy of all the baseline models from our training runs closely matched those reported in official PyTorch documentation\footnote{https://pytorch.org/docs/stable/torchvision/models.html}, shown in the last column in Table~\ref{tab:best} under PT\_official. 

It can be seen that partial convolution based padding (*\_partial) provides better validation accuracy than the default zero padding (*\_zero) for all the models. The network of VGG19 with partial convolution based padding has the largest improvement with 0.68\% accuracy boost. For the ResNet family, ResNet50 model has the largest improvement (0.478\%) compared with ResNet101 (0.36\%) and ResNet152 (0.248\%). This may be contrary to our straightforward intuition that deeper networks would benefit more from our proposed padding scheme, which is reflected in the comparison between VGG19 (0.687\%) and VGG16 (0.316\%) to some extent. On the other hand, this can also be interpreted as shallower networks have more potential and room for improvement. Furthermore, we found that reflection padding or replication padding leads to similar accuracies, shown from the $2$nd row to the $5$th row in Table~\ref{tab:best}. We will show the reflection padding and replication padding results of other networks in the supplementary file. 

From Tables~\ref{tab:best} and~\ref{tab:avg}, it can also been seen that models with partial convolution based padding have smaller standard deviation than the ones with zero padding. It means that partial convolution based padding makes model's performance stable and robust to randomness.

\begin{figure}
    \centering
    \includegraphics[width=\linewidth]{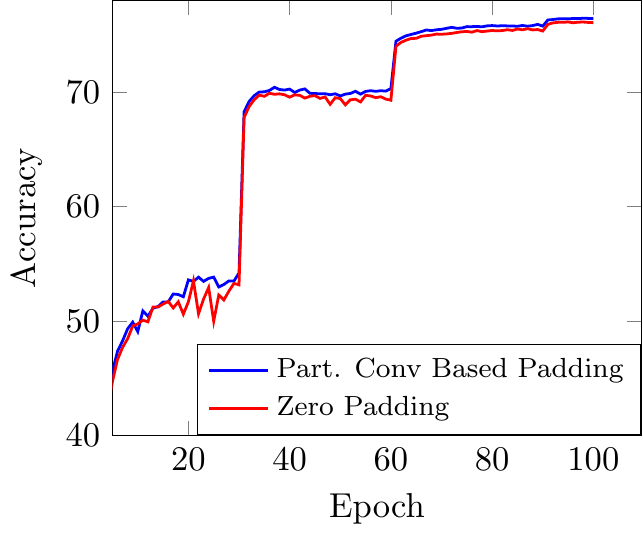}
    \caption{This plot shows the comparison of testing/validation accuracy between ResNet50 with partial convolution based padding and ResNet50 with zero padding from the model checkpoints at each epoch. ResNet50 with partial convolution based padding vs. zero padding. From the plot, it can be seen that to achieve the same accuracy partial convolution based padding takes fewer epochs; with the same training epoch, the model partial convolution based padding consistently achieves better accuracies than the one with zero padding.}
    \label{fig:rn50_convergence}
\end{figure}

\textbf{Convergence Speed}. Besides obtaining better accuracy, the model with partial convolution based padding can also achieve the similar/same accuracy with fewer iterations/epochs compared to the model with zero padding. In Figure~\ref{fig:rn50_convergence}, we plot the average testing/validation accuracy for every epoch's model checkpoint among the 5 runs of both ResNet50 with partial convolution based padding and ResNet50 with zero padding. Blue lines show the accuracy of the models with partial convolution based padding; yellow lines show the accuracy of the models with zero padding. It can be seen that: 

\begin{enumerate}
    \item The model with partial convolution based padding takes fewer iterations/epochs to achieve the similar/same accuracy; for example, to achieve the accuracy at $74.9\%$; the model with partial convolution based padding takes 63 epochs to achieve the score while the model with zero padding takes 68 epochs.
    \item  The training of the model with partial convolution based padding is more stable with fewer drifts compared to the model with zero padding shown as the blue and yellow lines in Figure~\ref{fig:rn50_convergence}.
\end{enumerate}

\begin{figure*}
    \centering
    \includegraphics[width=0.15\textwidth]{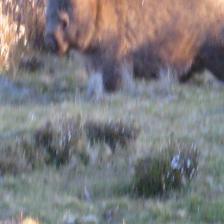}
    \includegraphics[width=0.15\textwidth]{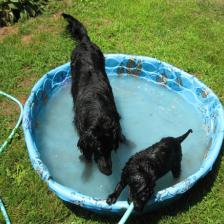}
    \includegraphics[width=0.15\textwidth]{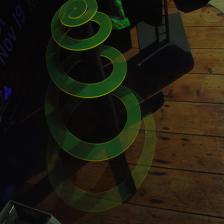}
    \includegraphics[width=0.15\textwidth]{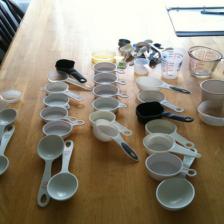}
    \includegraphics[width=0.15\textwidth]{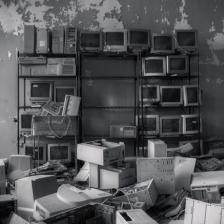} \\
    \includegraphics[width=0.15\textwidth]{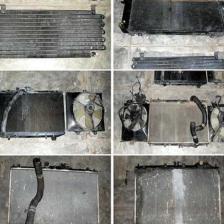}
    \includegraphics[width=0.15\textwidth]{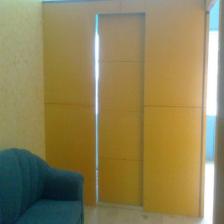}
    \includegraphics[width=0.15\textwidth]{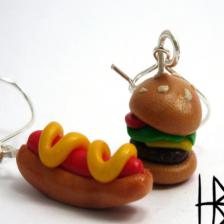}
    \includegraphics[width=0.15\textwidth]{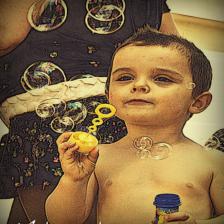}
    \includegraphics[width=0.15\textwidth]{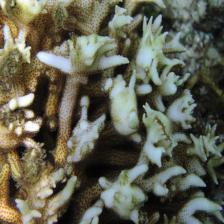}
    \caption{Images that fail at all the 5 runs of resnet50 with zero padding, but succeeded for all the 5 runs with partial convolution based padding.}
    \label{fig:zero_failures}
\end{figure*}

\begin{figure*}
\centering
{\includegraphics[width=0.8\textwidth]{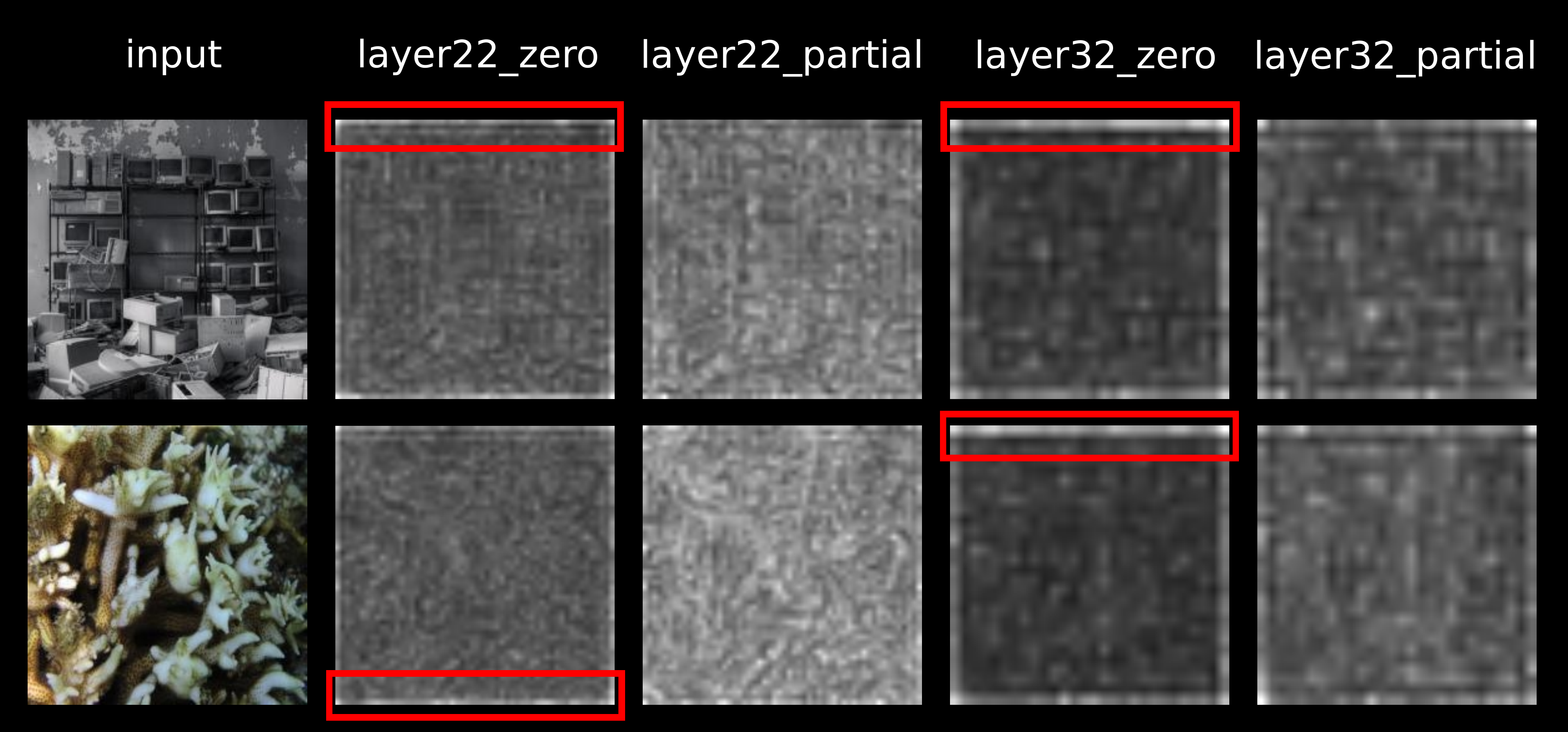}}
\caption{Activation Map at 22th layer and 32th layer of VGG19 network with zero paddding and VGG19 network with partial convolution based padding. These two layers are ReLU layers and we sum up the activation along channels and resize the summation for visualization. *\_zero shows the activations from VGG19 network with zero padding; *\_partial shows the activations from the partial convolution based padding version. Red rectangles show the strong activation regions from VGG19 network with zero paddding.}
\label{fig:activation_map}
\end{figure*}

\textbf{Activation Map Visualization}. Figure~\ref{fig:zero_failures} shows the image examples whose predictions fail for all the 5 runs of resnet50 with zero padding, but succeed for all the 5 runs of resnet50 with partial convolution based padding. The same activation maps of the last column's two images are shown in Figure~\ref{fig:activation_map}. It can be seen that for the zero padding based network, the features at the border have the strongest activation responses, which easily confused the network prediction. As these border features are largely dependent on the padded zero values from the previous layers, this could to some extent also imply that there is a chance that the padded zero values could mislead the network.

\begin{table*}[h]
    \centering
    \begin{tabular}{cc|ccccc|cc}
    Training & Inference &1&2&3&4&5&average&diff \\
    \hline
    resnet50\_zero & resnet50\_zero&76.240&76.074&75.988&76.136&76.224&76.132& - \\
    resnet50\_zero & resnet50\_partial& 62.364 & 55.576 & 56.746 & 61.824 & 62.364 & 59.7748 & -16.357\\
    \hline
    resnet50\_partial&resnet50\_partial&76.606&76.532&76.638&76.562&76.716&76.611& - \\
    resnet50\_partial&resnet50\_zero&76.054 & 75.862 & 75.806 & 75.854 & 75.820 & 75.879 & -0.732 \\
    \end{tabular}
    \caption{Cross testing of using one padding scheme for training and the other padding scheme for inference. It can be seen that if the training is with zero padding but inference is with partial convolution based padding, the accuracy drop is much bigger than the opposite direction.}
    \label{tab:cross_testing}
\end{table*}

\textbf{Cross Testing}. One feature of partial convolution based padding is that it does not introduce extra parameters. Thus, it allows us to perform such cross testing that we use the network weights trained with one padding scheme to do the inference with the other padding schemes. Table~\ref{tab:cross_testing} shows such cross testing on resnet50\_zero and resnet50\_partial (resnet50\_zero with partial convolution based padding); it can be seen that if the training is using partial convolution based padding while the inference is using zero padding, there is only a $0.732\%$ accuracy drop; to some extent, it is even comparable (close) to the  accuracy obtained by using zero padding for both training and testing. However, if the training is using zero padding and the inference is using partial convolution based padding, there would be a big accuracy drop with $16.357\%$. One possible explanation could be that the model weights trained with zero padding are sensitive to the image/feature value (scale) changes at the image/feature border; on the other hand, the model trained with partial convolution based padding is less sensitive to such changes. It may also be interpreted as models with zero padding have paid more efforts to resolve the influence brought by adding zero values at the border.

\subsection{Semantic Segmentation Network}

\begin{table*}[h]
    \centering
    \begin{tabular}{l|c|c|c|c|c||c|c|c|c|c}
    \multicolumn{1}{c|}{} & \multicolumn{5}{c||}{default split} & \multicolumn{5}{c}{additional split} \\
     & 1\_best & 2\_best & 3\_best & mean & diff & 1\_best & 2\_best & 3\_best & mean & diff \\
    \hline
    RN50\_zero & 78.025 & 78.081 & 78.146 & 78.084 & - & 77.06 & 76.249 & 76.44 & 76.583 \\
    RN50\_partial & 78.372 & 78.006 & 78.235 & 78.204 & 0.120 & 76.751 & 76.955 & 77.031 & 76.912 & 0.329 \\
    \hline
    \hline
    WN38\_zero & 80.163 & 80.042 & 80.397 & 80.201 & - & 79.069 & 78.743 & 78.707 & 78.840 & - \\
    WN38\_partial & 80.482 & 80.357 & 80.101 & 80.313 & 0.112 & 79.045 & 78.885 & 79.082 & 79.004 & 0.164
    \end{tabular}
    \caption{DeepLabV3+ evaluation mIOU($\%$) difference on CityScapes dataset. *\_zero and *\_partial indicate  the corresponding model with zero padding and partial convolution based padding respectively. Both models are trained from the scratch on the training set of CityScape dataset and evaluated on the validation set of CityScapes dataset.}
    \label{tab:seg}
\end{table*}

Semantic segmentation involves densely classifying each pixel with its corresponding semantic category. Most recent semantic segmentation networks use an encoder-decoder architecture~\cite{ronneberger2015u,badrinarayanan2015segnet} to ensure the output dimensions match those of the input. It is common to employ an ImageNet pretrained classifier such as ResNet50 as the backbone for the encoder part, followed by a series of deconvolutions or upsampling layers for the decoder. Padding is essential in guaranteeing same input-output dimensions as the center of the filter would not be able to reach the edge pixels otherwise.

Some networks also use dilated convolutions to achieve larger receptive fields without downsampling or increasing the number of filter parameters. For dilated/atrous convolution, large padding is usually needed, increasing the dependence on padded values. Our partial convolution based padding formulation is general and is easily adapted to dilated convolutions.

Dilated convolutions are used in DeepLabV3+\cite{chen2018encoder}, one of the state-of-the-art semantic segmentation networks. DeepLabV3+ uses pretrained ImageNet classifier like Xception~\cite{szegedy2015going} or ResNet as the encoder backbone. The backbone features are fed into an dilated-convolution-based Atrous spatial pyramid pooling module (ASPP) to complete the encoding. Next, a decoder with the skip links to the encoder features and final upsampling is used to upscale the output to the original input size. We train and test a DeepLabV3+ model on the CityScapes semantic segmentation dataset. CityScapes contains $5000$ images with pixel-level annotations. The default splits are $2975$ for the training set, $500$ for the validation set and $1525$ for the test set. It also contains $20000$ coarsely annotated images. 

\begin{figure*}
\centering
\scalebox{0.9}{
\begin{tabular}{cccc}
\multicolumn{4}{c}{}\\
Input & G.T. Segmentation & zero padding & partial conv based padding \\
    \includegraphics[width=0.25\textwidth]{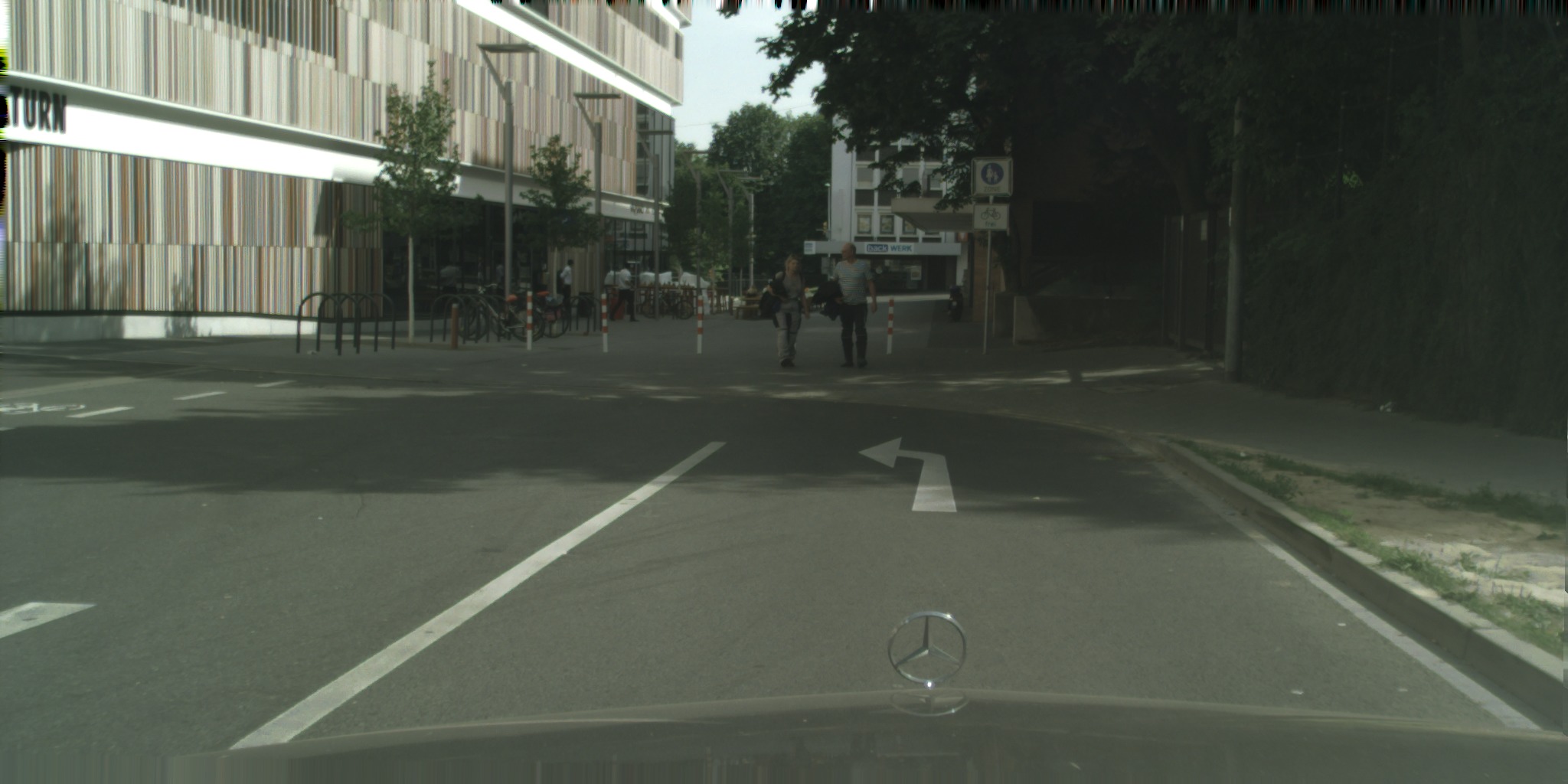}&
    \includegraphics[width=0.25\textwidth]{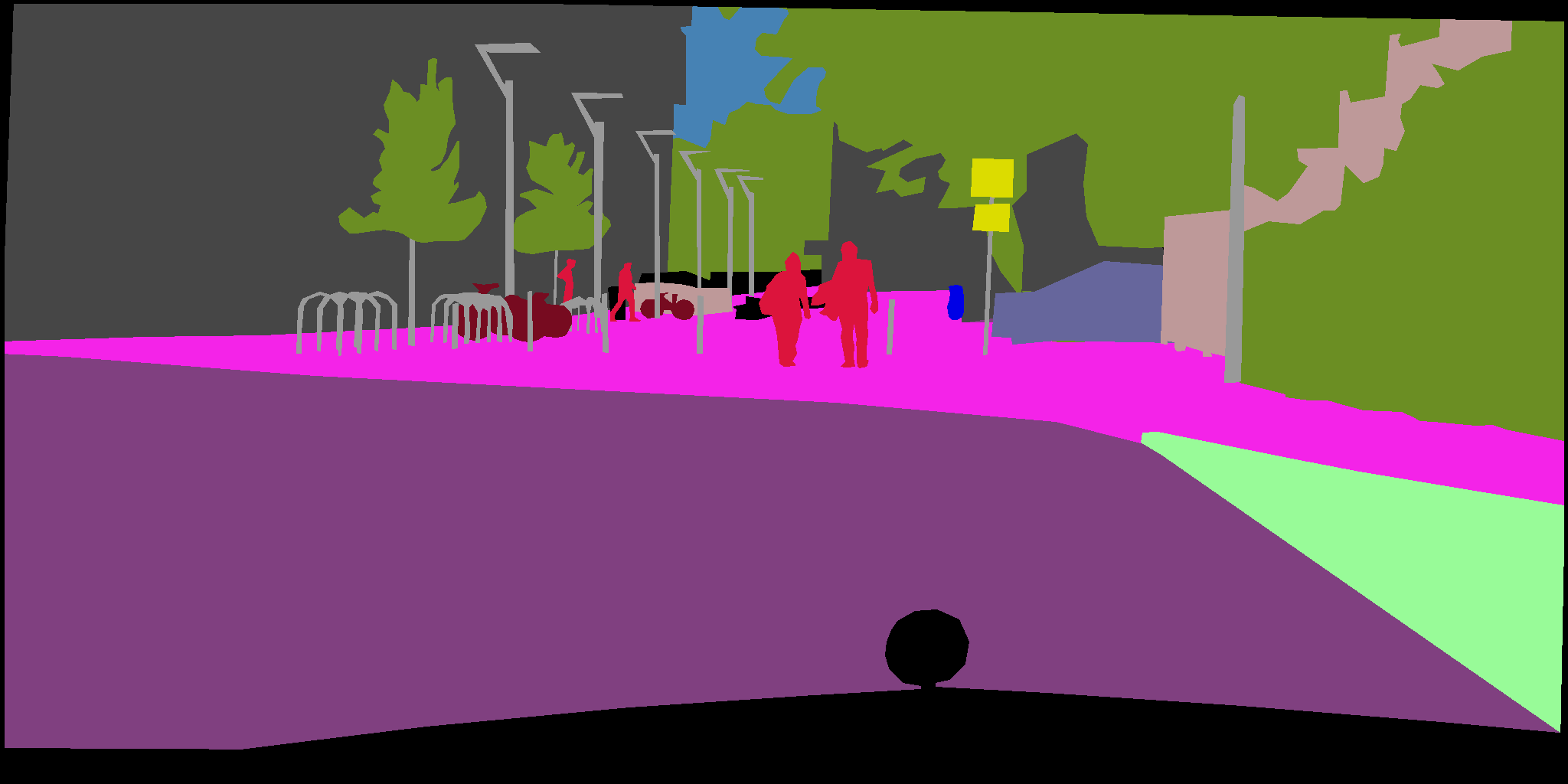}&
    \includegraphics[width=0.25\textwidth]{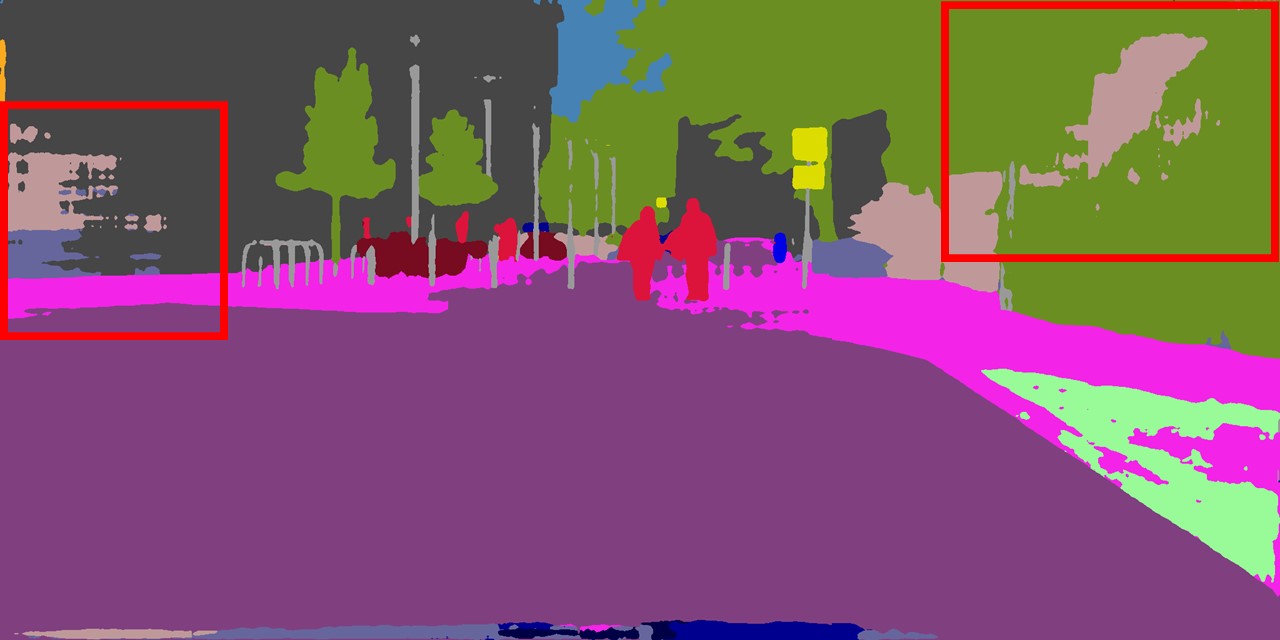}&
    \includegraphics[width=0.25\textwidth]{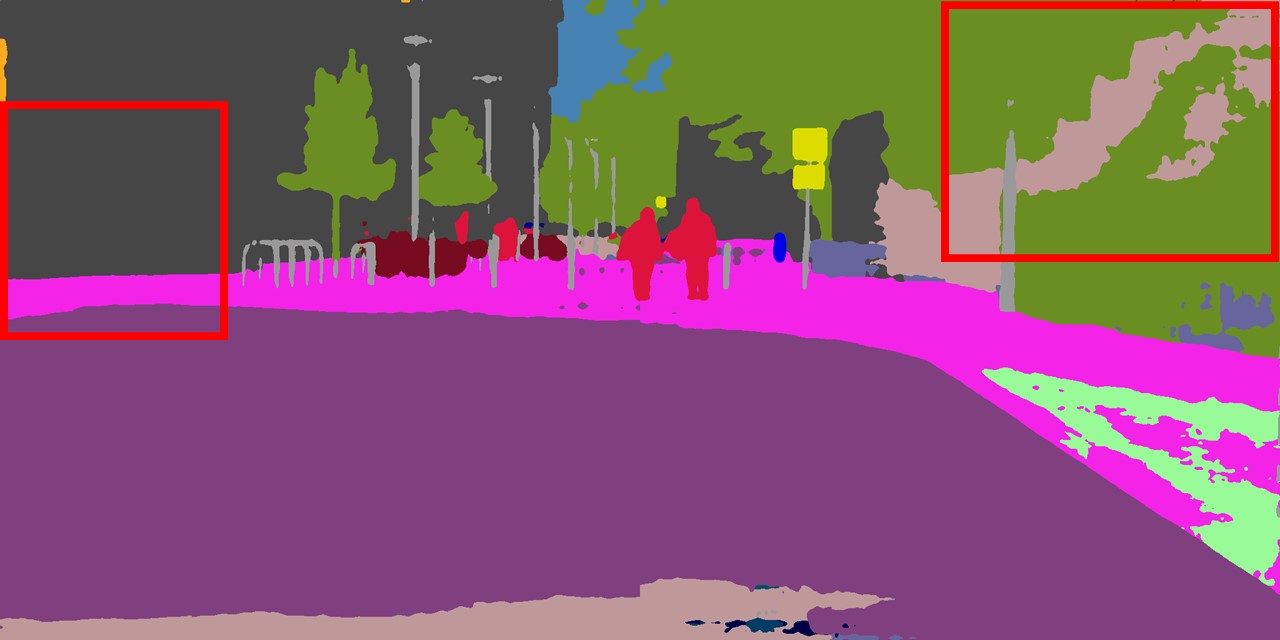}\\
    \includegraphics[width=0.25\textwidth]{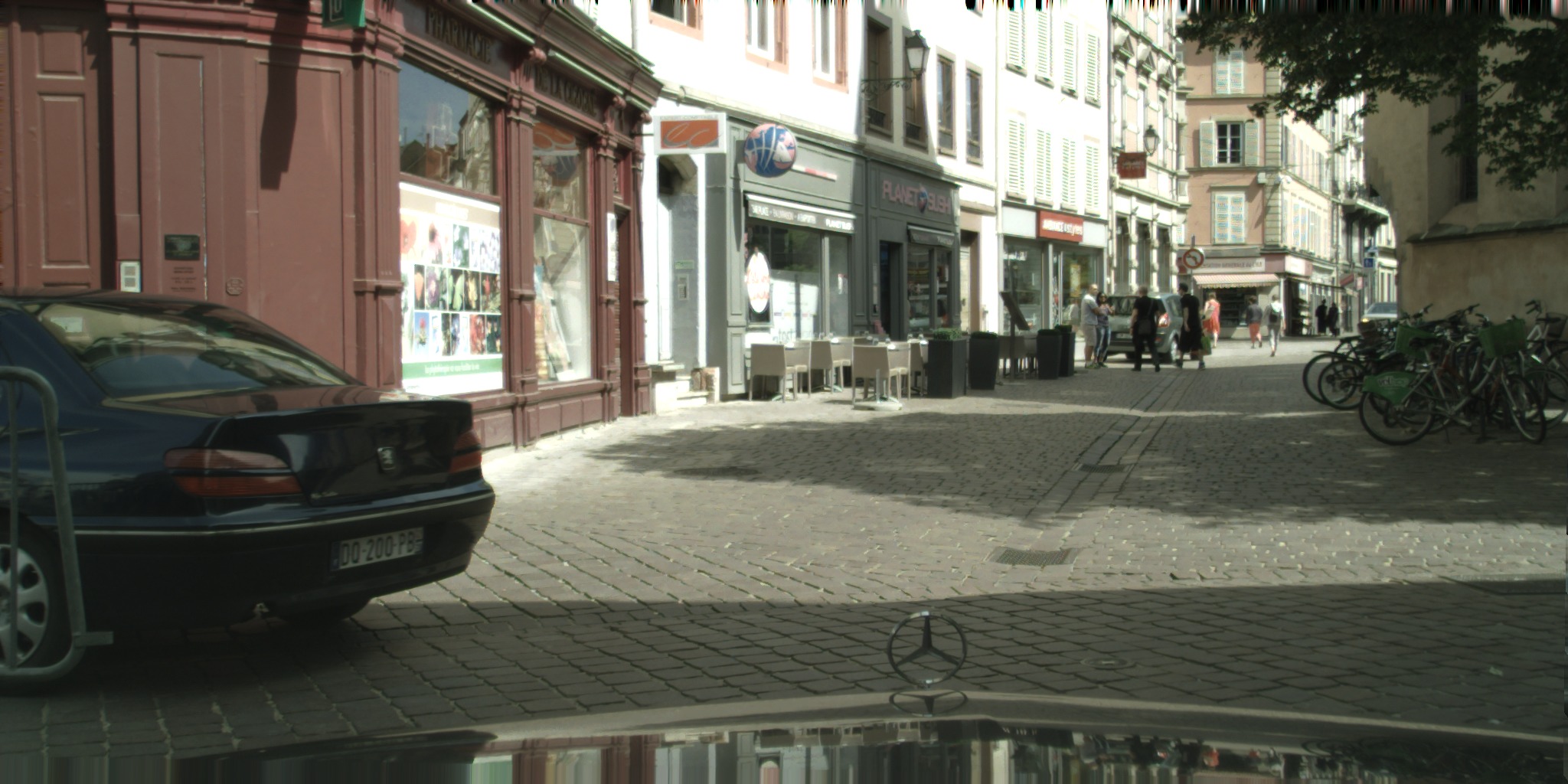}&
    \includegraphics[width=0.25\textwidth]{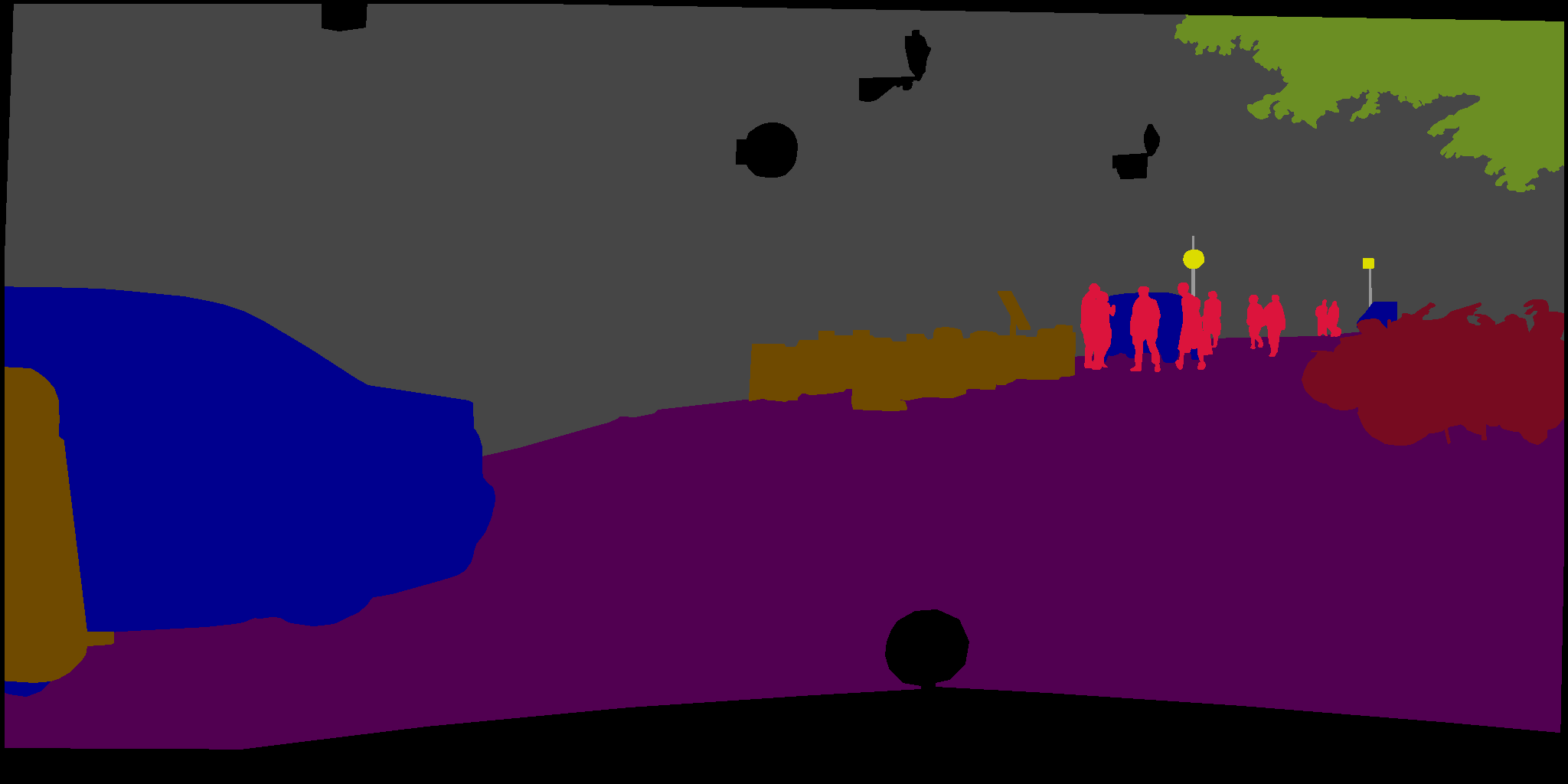}&
    \includegraphics[width=0.25\textwidth]{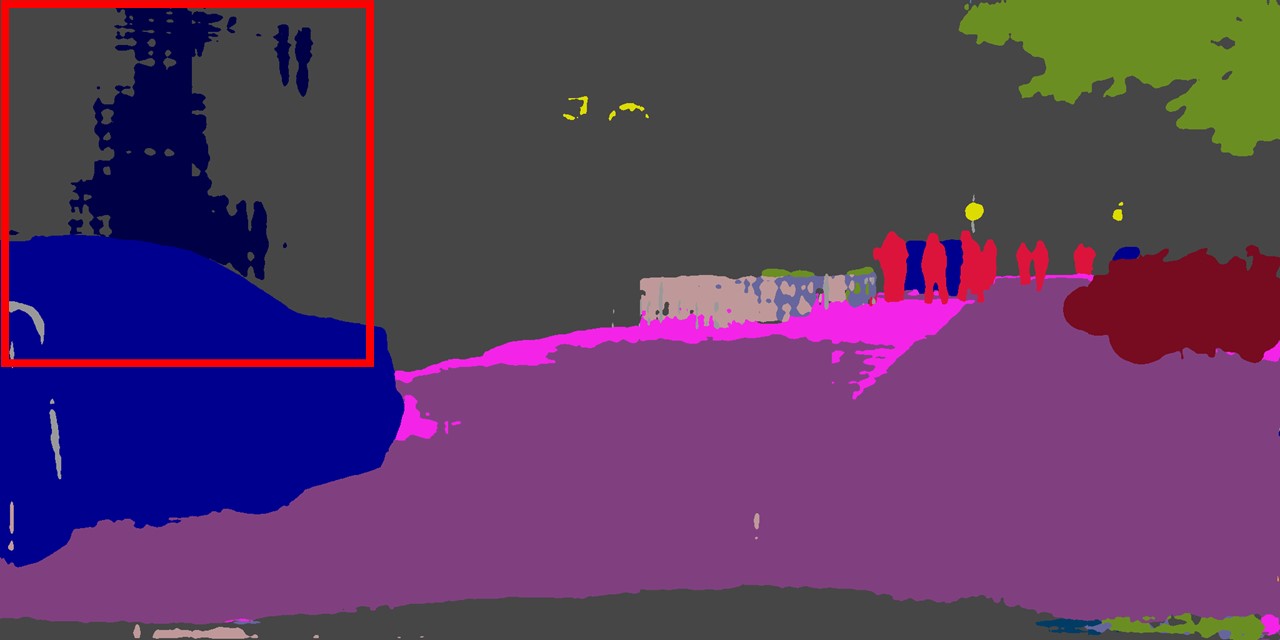}&
    \includegraphics[width=0.25\textwidth]{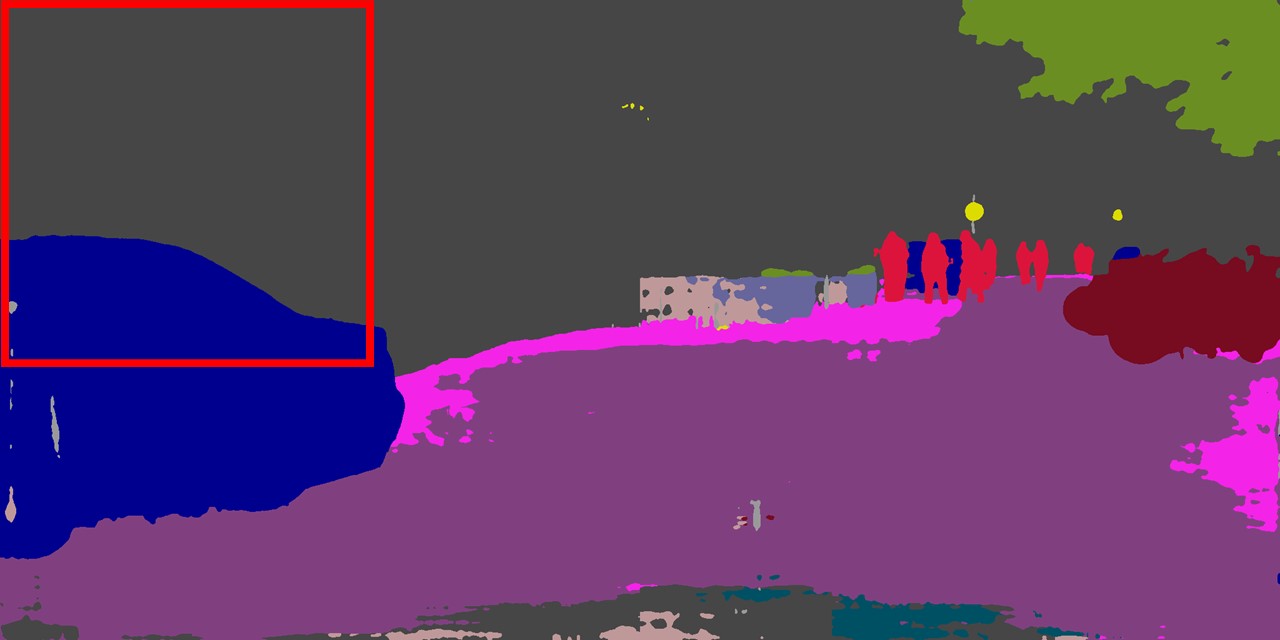}\\
    \includegraphics[width=0.25\textwidth]{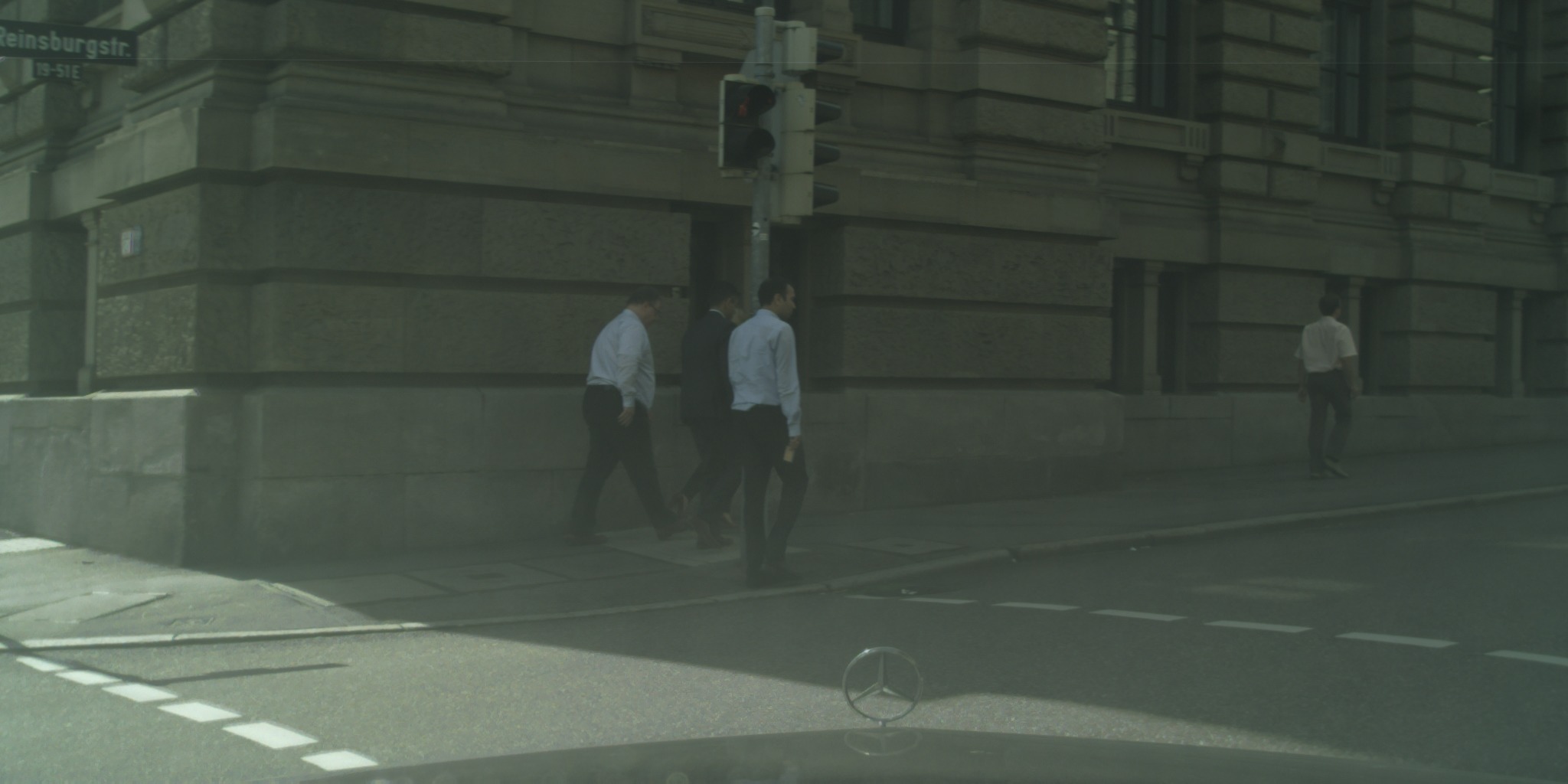}&
    \includegraphics[width=0.25\textwidth]{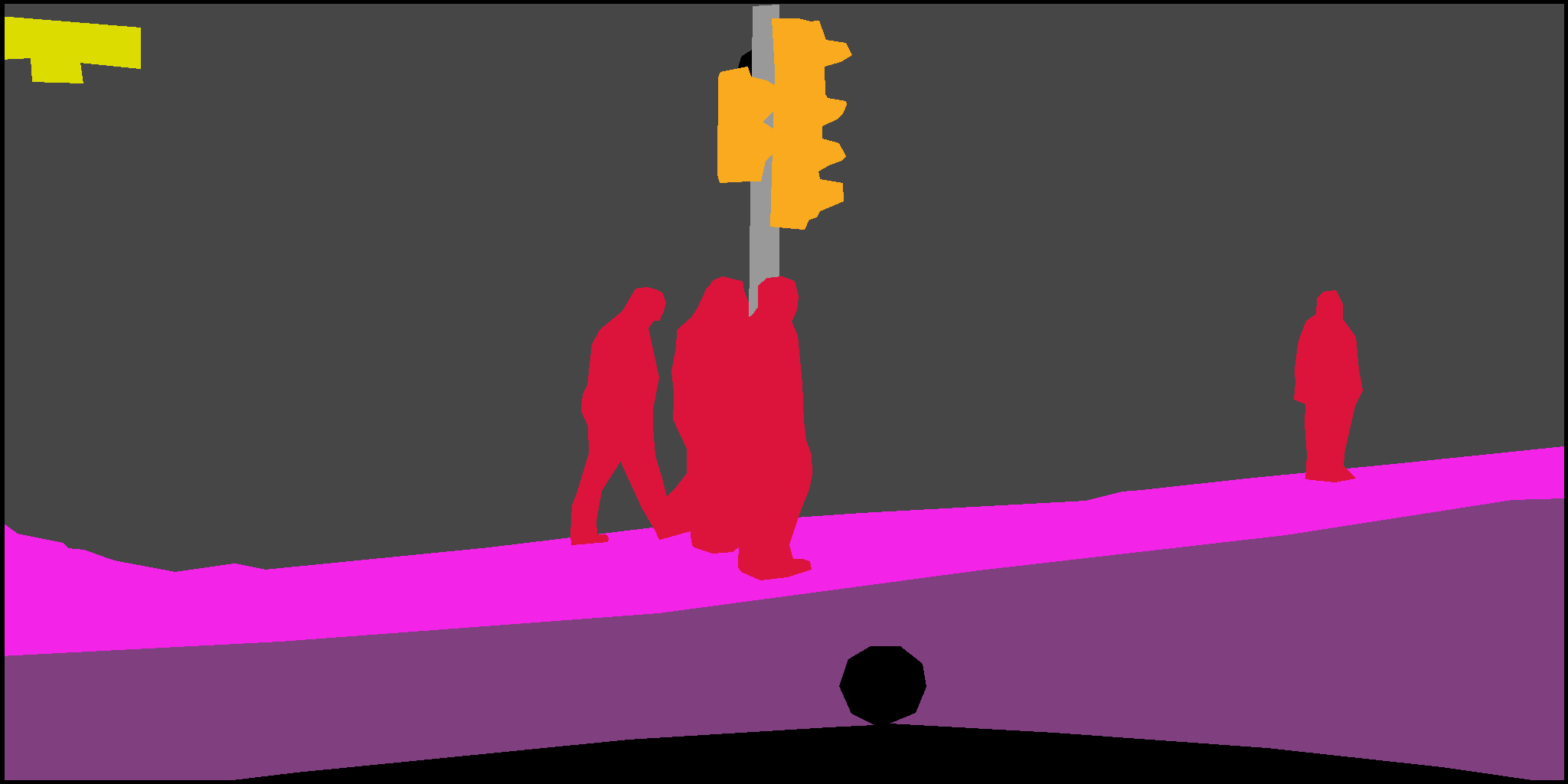}&
    \includegraphics[width=0.25\textwidth]{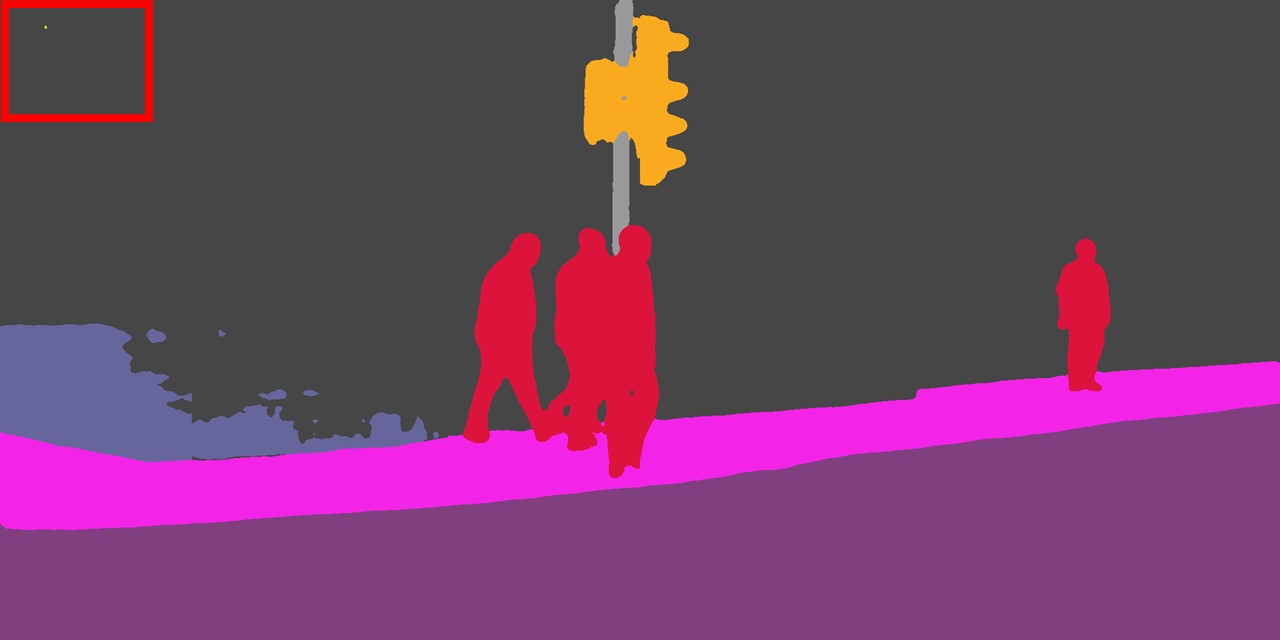}&
    \includegraphics[width=0.25\textwidth]{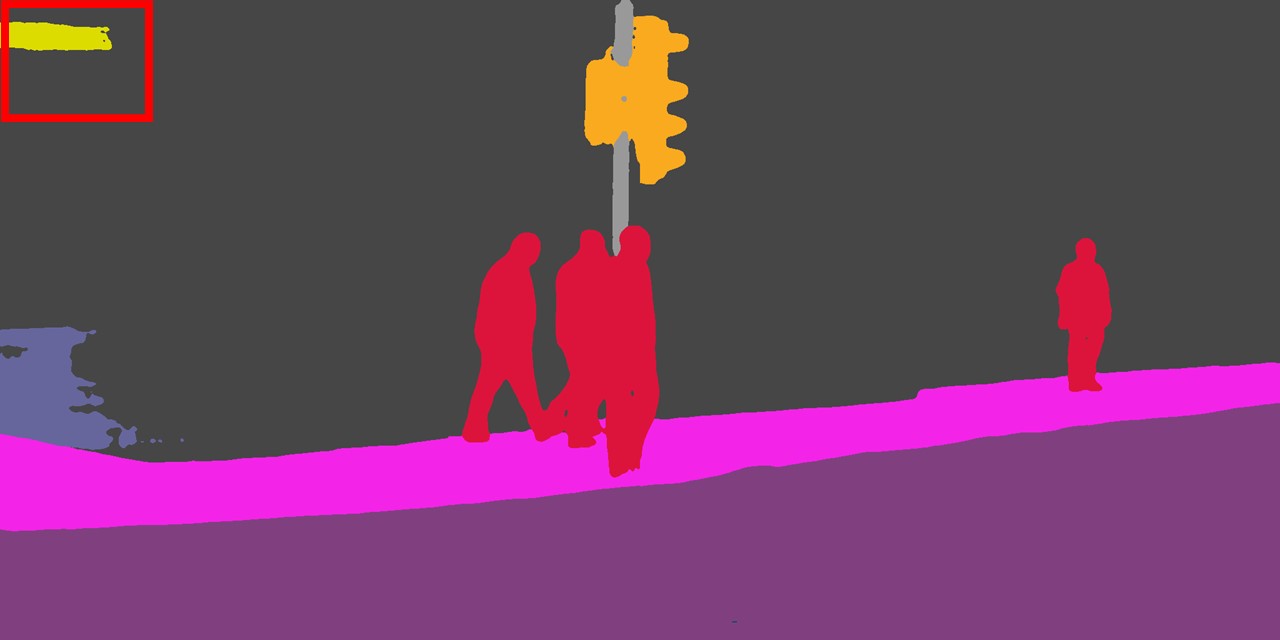}\\   
\end{tabular}
}
\caption{Semantic segmentation using tile based evaluation with $\frac{1}{3}$ tile overlapping: Visual comparison on Cityscapes.  From left to right:  Image,  Ground Truth Segmentation, zero padding prediction, partial conv based padding prediction. We demonstrate that partial convolution based padding method can remove border  artifacts thus resulting in a better prediction.}
\label{fig:seg_result_0.3overlap}
\vspace{-.3cm}
\end{figure*}

\begin{table}
    \centering
    \begin{tabular}{c|c|cc}
        Padding & Regular & 0 Tile overlap & $\frac{1}{3}$ Tile overlap\\
        \hline
        WN38\_zero & 79.069 & 79.530 & 79.705 \\
        WN38\_partial & 79.082 & 79.907 & 80.079 \\
    \end{tabular}
    \caption{Evaluation: Comparing Tile based evaluation vs Regular Full image evaluation. *\_zero and *\_partial, indicate  the corresponding model with zero padding and partial convolution based padding respectively}
    \label{tab:eval_overlap}
    \vspace{-.3cm}    
\end{table}



To keep our experimentation simple and focused on the differences between regular and partial convolution based padding, we do not employ external datasets for pre-training as is done to achieve state-of-the-art performance in works such as~\cite{bulo2017place}.

The Cityscapes dataset contains training data from $21$ different cities. The default train-val split creates an $18$/$3$ train/val split by cities. We create an additional second $18$/$3$ split to experiment on as well. Our segmentation network architecture is based on DeepLabV3+~\cite{chen2018encoder} with output stride of $8$ for the encoder. Motivated by Mapillary~\cite{bulo2017place}, we evaluate partial convolution based padding using WideResnet38~\cite{wu2016wider} and Resnet50 for our encoder. We also use a data sampling strategy similar to~\cite{bulo2017place} and use the 20k coarsely annotated images along with the finely annotated images. We run the segmentation network for 31K iterations with SGD with an initial learning rate of $3e-2$ and $1e-2$ for Resnet50 and WideResnet38 respectively and with a polynomial learning rate decay of $1.0$. Our momentum and weight decay are set to $0.9$ and $1e-4$ respectively. Similar to other semantic segmentation papers ~\cite{zhao2017pyramid,zhang2018context,chen2018encoder}, we use the following augmentations during training: random scaling, horizontal flipping, Gaussian blur, and color jitter. Our crop size was set to 896 and 736 for Resnet50 and WideResnet38 respectively. Lastly to due to the large crop size, we use a batch size of $2$ with synchronized BatchNorm (for distributed training), similar to PSPNet~\cite{zhao2017pyramid}.

\textbf{Analysis}.
We compare and evaluate the segmentation models using mIOU metric ($\%$). The mIOU is the mean IOU across 19 classes for cityscapes. 
The ASPP module includes a spatial pooling step that outputs a single 1-D feature vector. During training, this pooling procedure operates on square crops (1:1 aspect ratio). However, during full-image inference, we must now account for the fact that the full images are of size 1024$\times$2048 with an aspect ratio of 1:2. This means that the pooling module, trained to pool over 1:1 aspect ratios, must now pool over 1:2 aspect ratio input at test time. We resolve this by breaking down the full image into overlapping square tiles of 1:1 aspect ratio. We report two types of results:
\begin{enumerate}
\item \textbf{regular:} directly feed the $1024 \times 2048$ image into the network regardless of the aspect ratio issue.
\item \textbf{tile:} dividing the images into square tiles of size $1024 \times 1024$.  
\end{enumerate}

Tile based evaluation is also used in the work~\cite{zhao2017pyramid} to obtain better evaluation performance. In Table~\ref{tab:seg}, we show that our segmentation models using partial convolution based padding with Resnet50 and WideResnet38 encoder backbones achieve better mIOU on full image (regular). Resnet50 encoder based segmentation model trained using partial convolution based padding achieved $0.12\%$ and $0.329\%$ higher mIOU on the default and additional split respectively. We also observe similar performance gains with WideResnet38+partial convolution based padding outperforming its counterpart by $0.112\%$ and $0.164\%$ in the default and additional split respectively.  

In Table~\ref{tab:eval_overlap}, we see that the tile based evaluation gives better mIOU than the regular evaluation even though their mIOUs on regulari evaluation mode are similar.

\textbf{Advantage of Partial Convolution based Padding for Tile base Evaluation:} One major concern for tile-based evaluation is that by subdividing the image, we significantly increase the number of pixels that lie near the boundaries of the input. As previously stated, this can hurt performance because pixels will have incomplete context when the receptive field extends beyond the image boundaries. We ameliorate this by sampling tiles with overlapping context -- allowing a boundary pixel in one tile to be re-sampled near the center in the neighboring tile. 

While overlapping tiles can serve to de-emphasize the boundary issue, we demonstrate in Table~\ref{tab:eval_overlap} that the partial convolution based padding models demonstrate a much larger improvement from tiling evaluation. This is because the latter type of model is more robust to the boundary issue in the first place, and thus was much less affected by the increase in pixels near borders. For both the evaluation with non overlapping between tiles and the evaluation with $\frac{1}{3}$ overlapping between tiles, the model with partial convolution based padding is around 0.37\% better than the model with zero padding, despite both having similar mIOUs in the regular evaluation scheme.

\begin{figure*}[h!]
\centering
\scalebox{0.9}{
\begin{tabular}{cccc}
\multicolumn{4}{c}{}\\
Input & G.T. Segmentation & zero padding & partial conv based padding \\
    \includegraphics[width=0.25\textwidth]{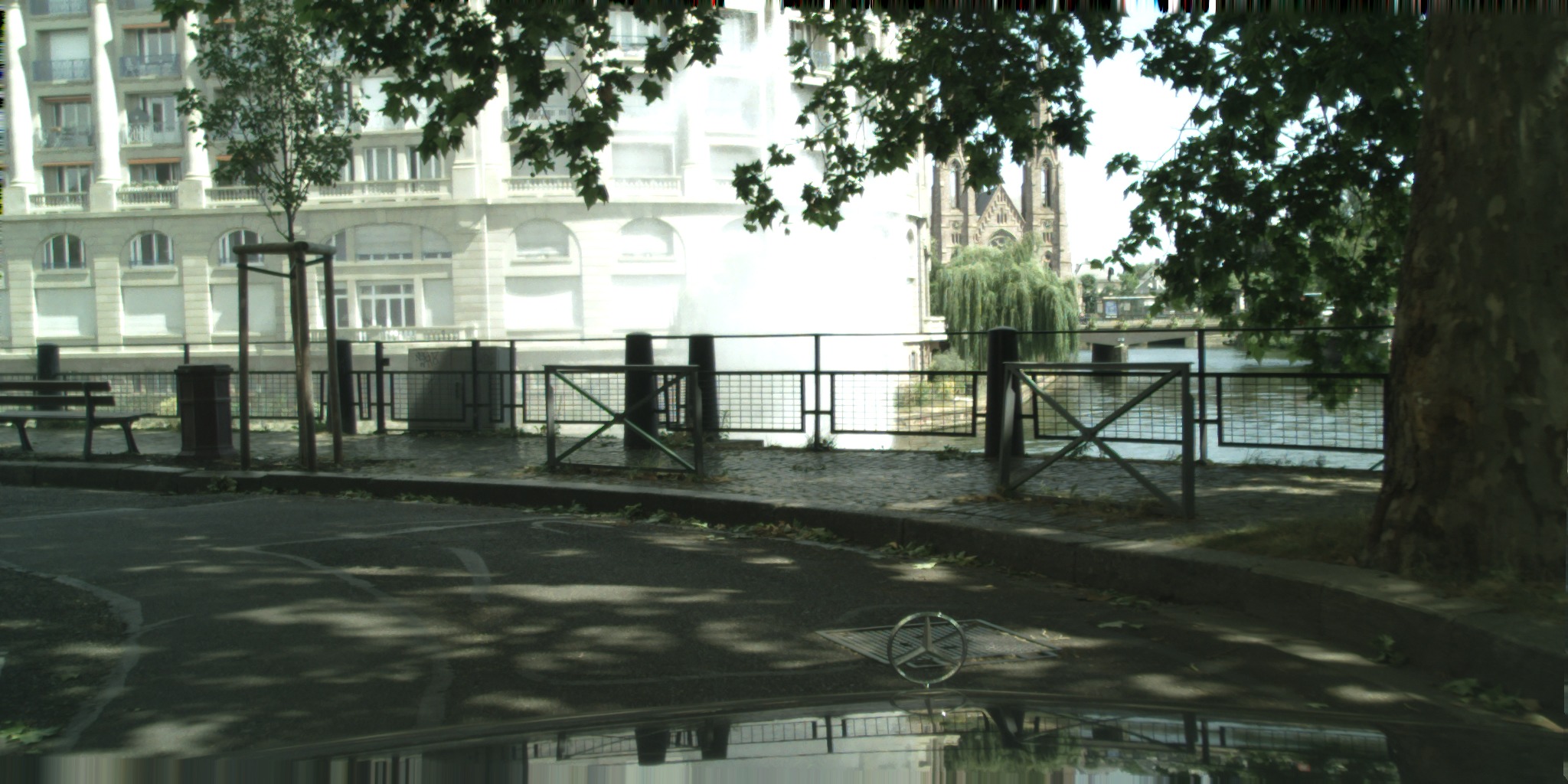} &
    \includegraphics[width=0.25\textwidth]{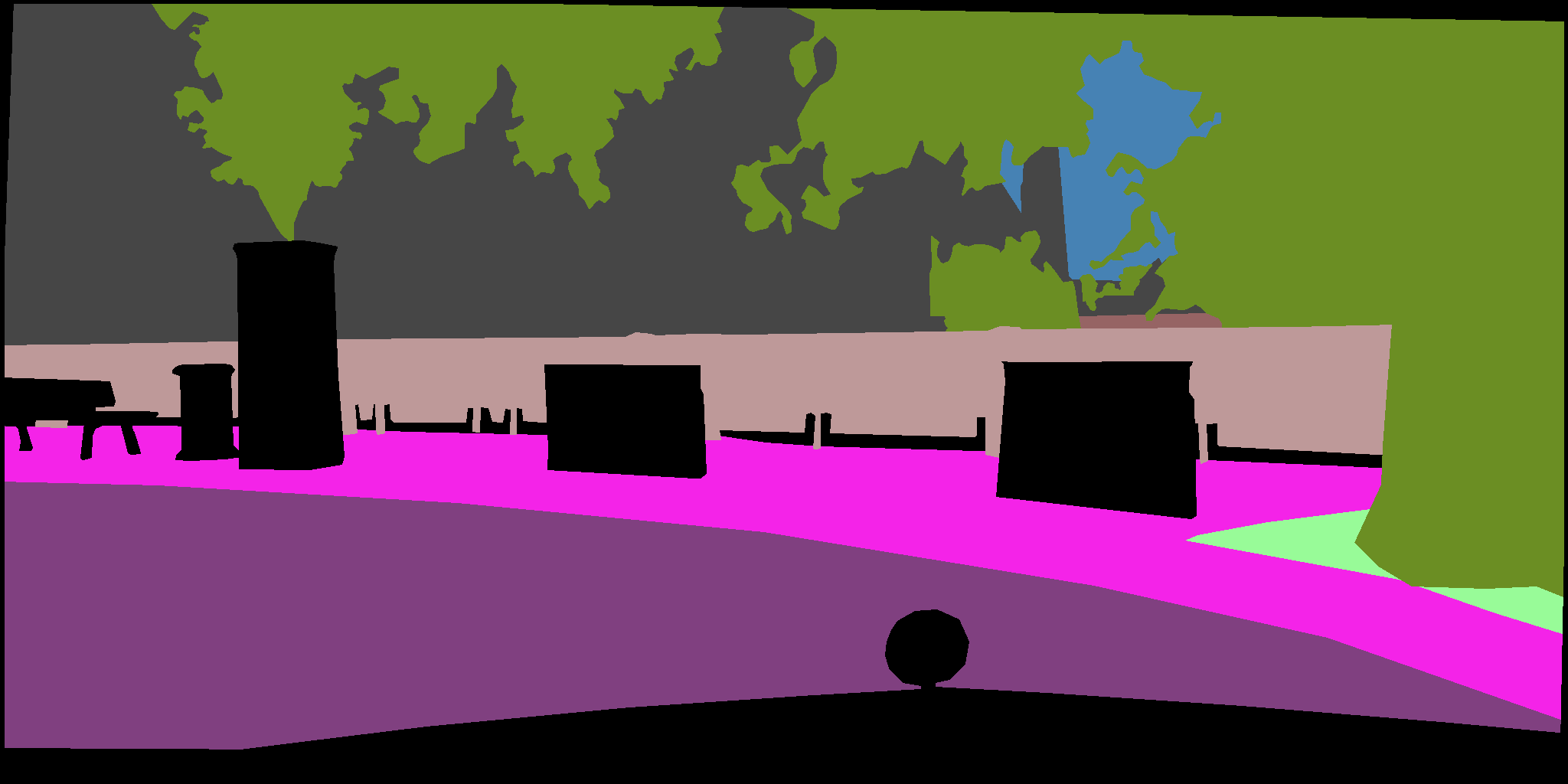} &
    \includegraphics[width=0.25\textwidth]{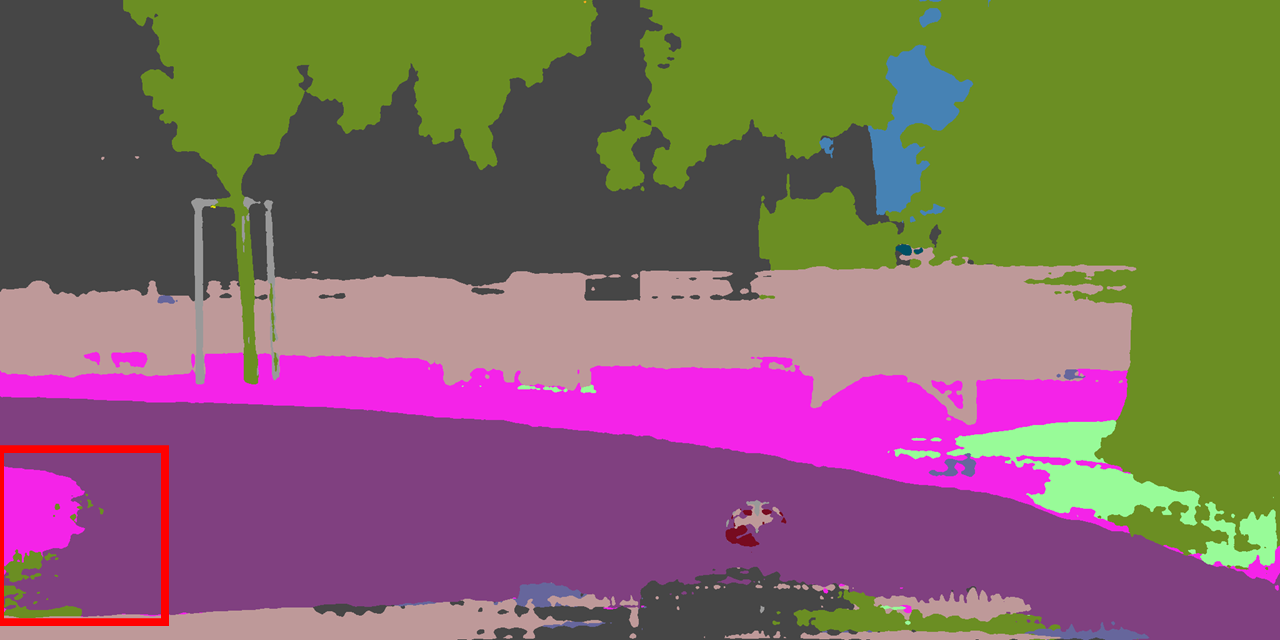} &
    \includegraphics[width=0.25\textwidth]{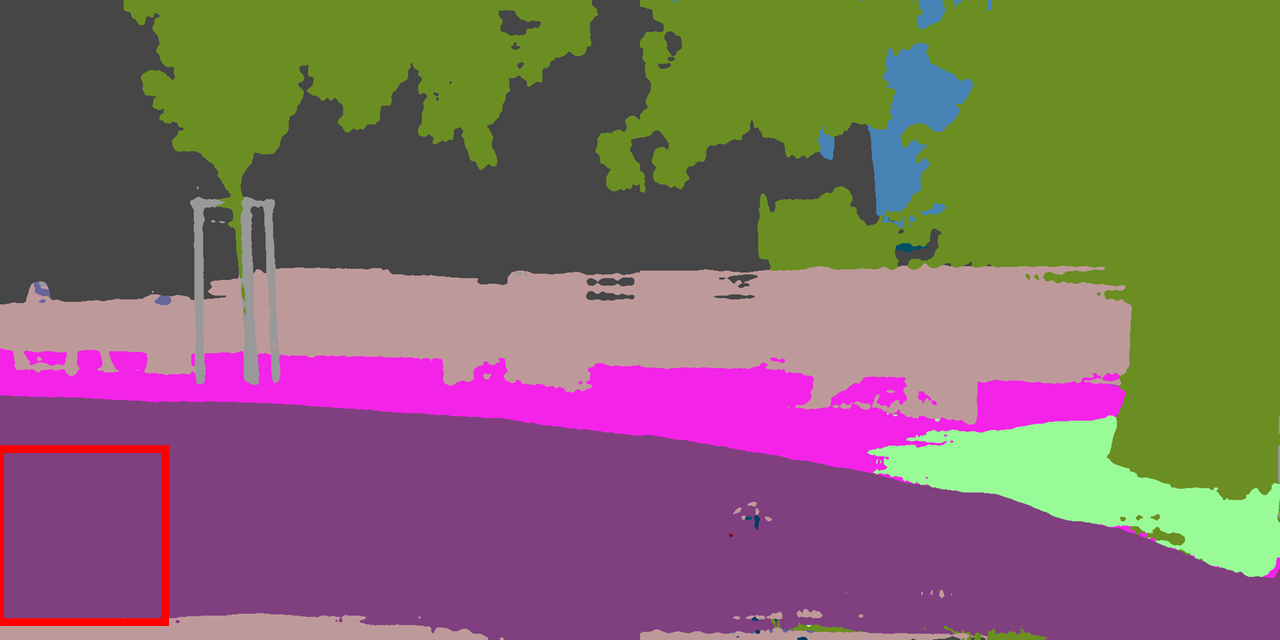} \\
    \includegraphics[width=0.25\textwidth]{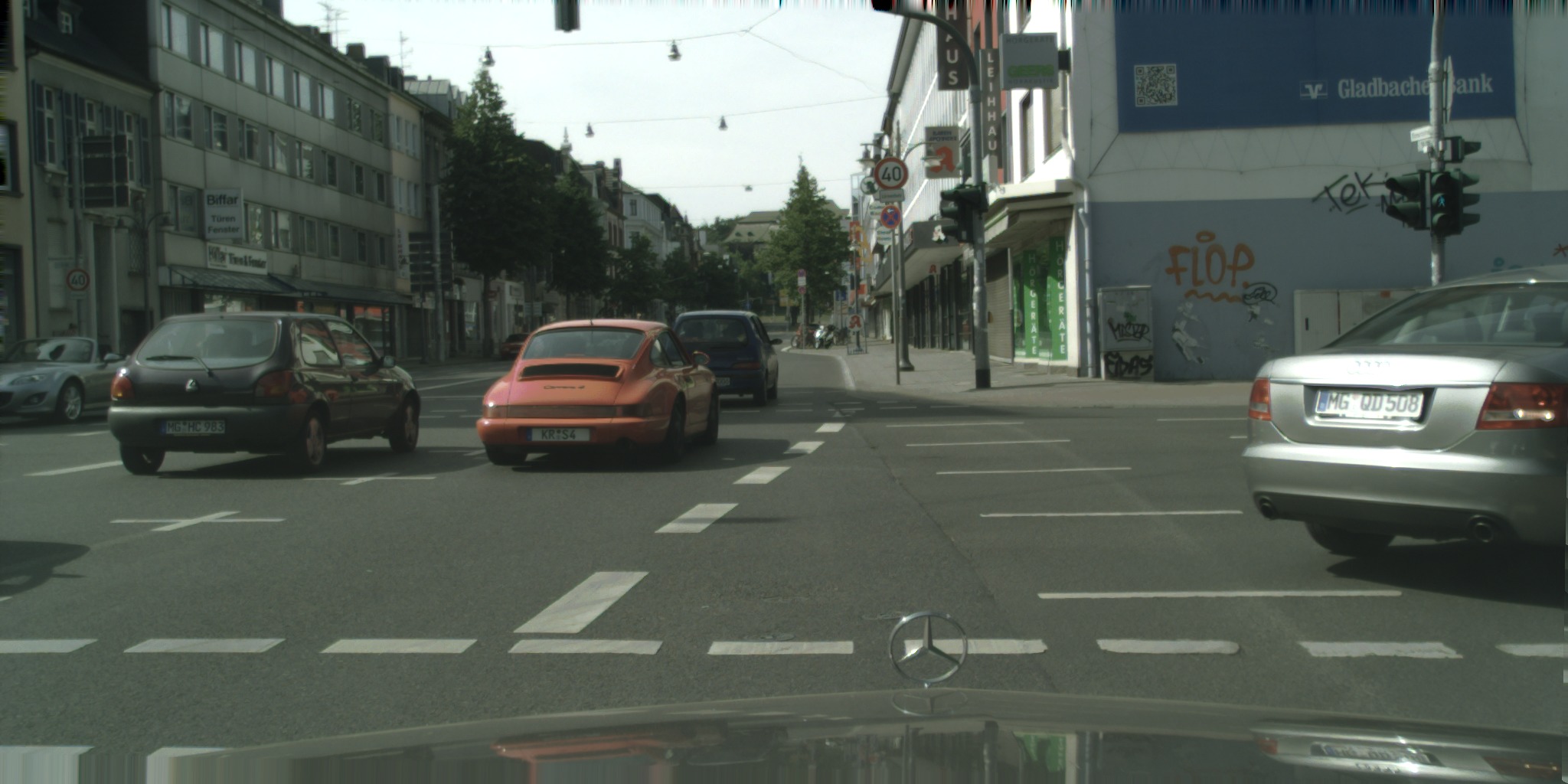}&
    \includegraphics[width=0.25\textwidth]{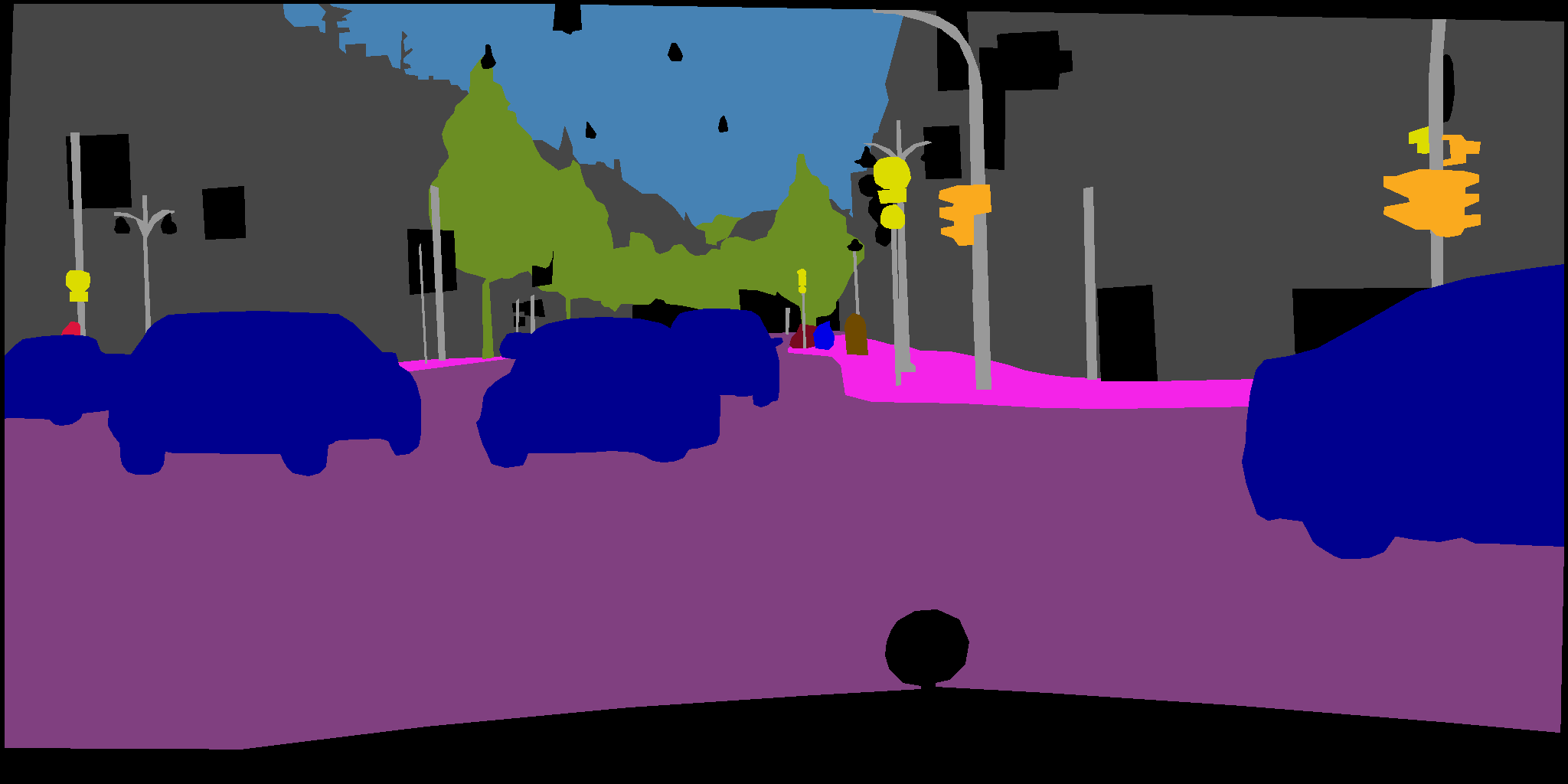}&
    \includegraphics[width=0.25\textwidth]{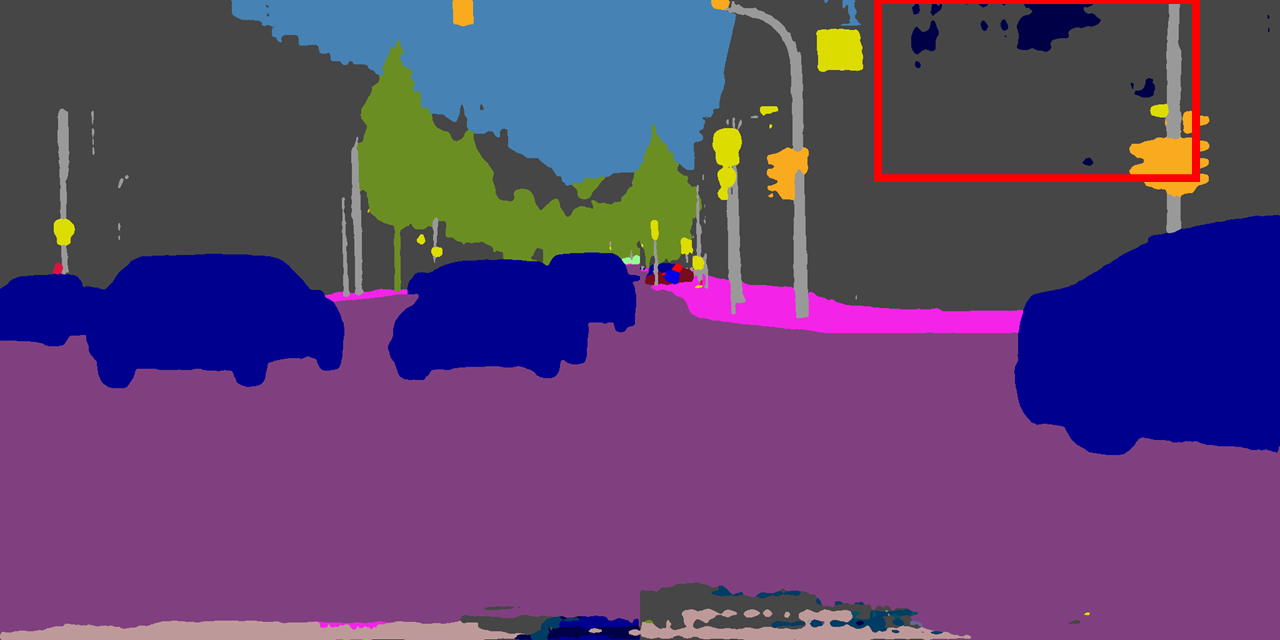}&
    \includegraphics[width=0.25\textwidth]{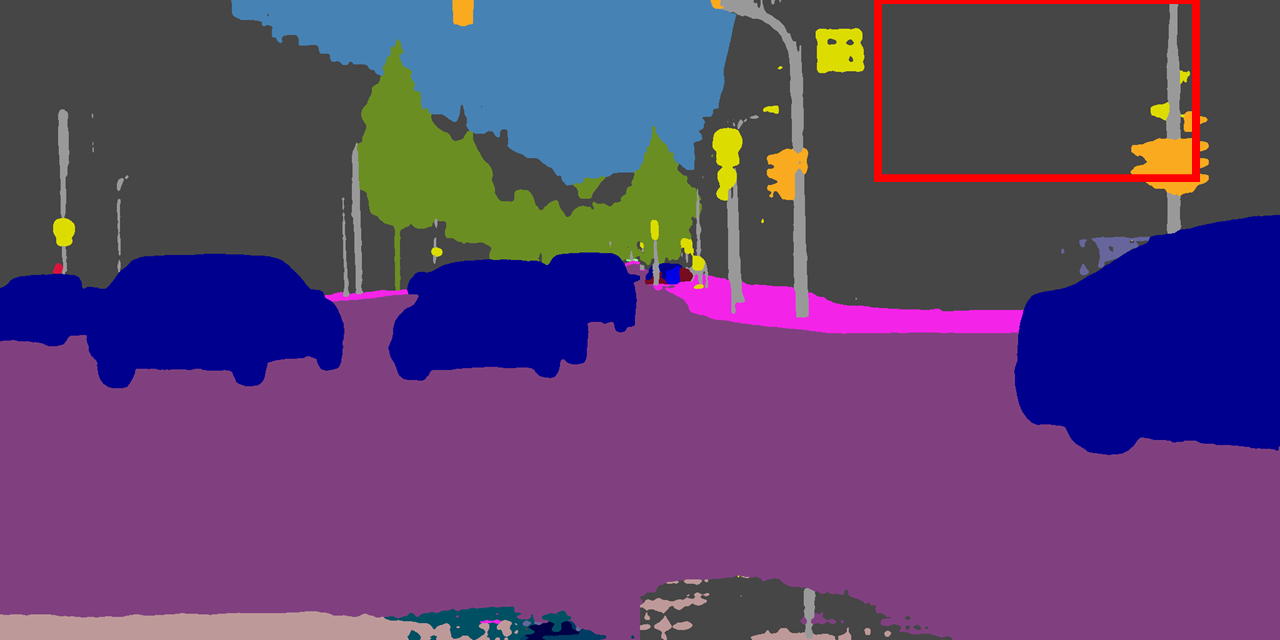} \\
    \includegraphics[width=0.25\textwidth]{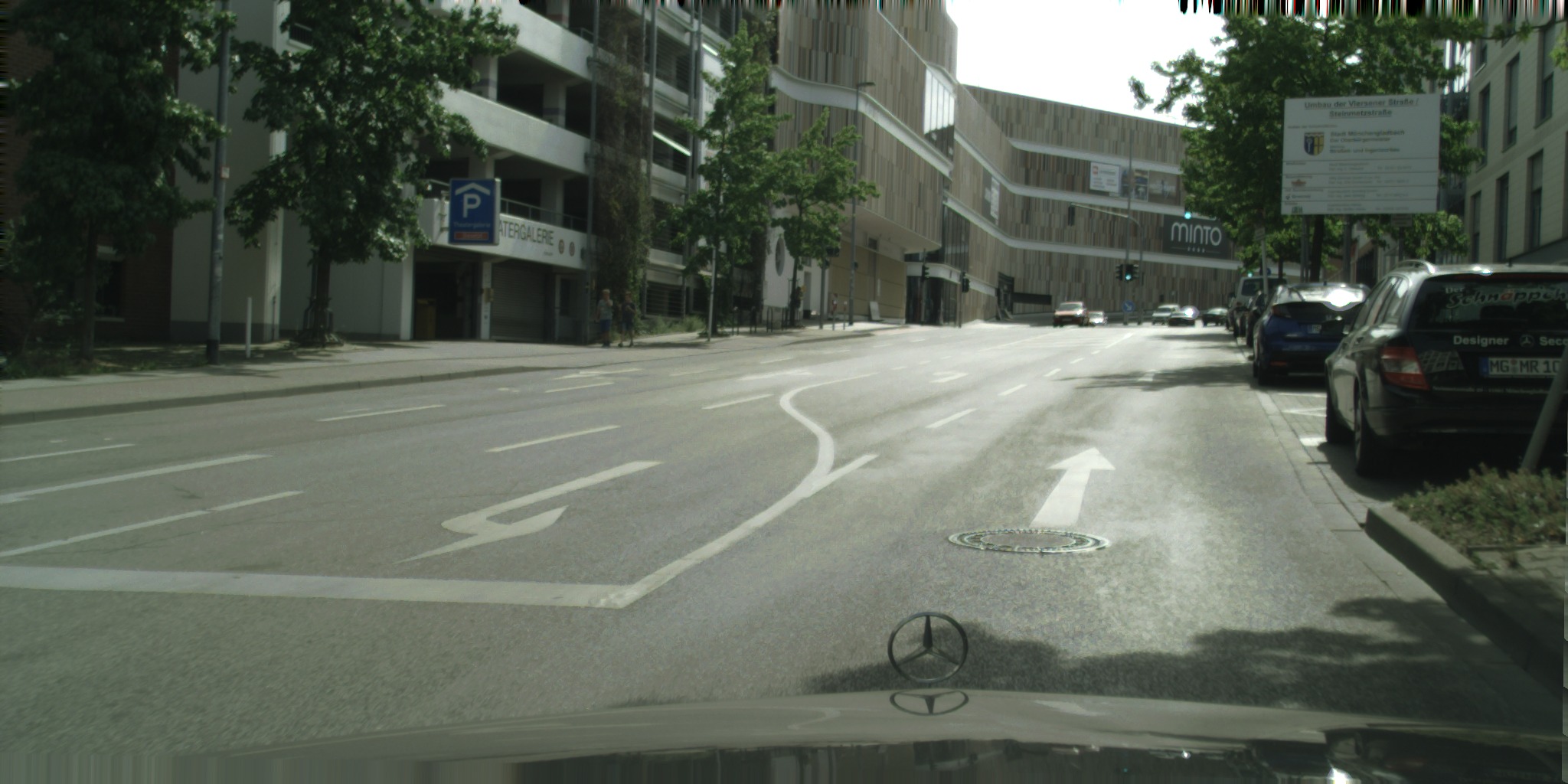} &
    \includegraphics[width=0.25\textwidth]{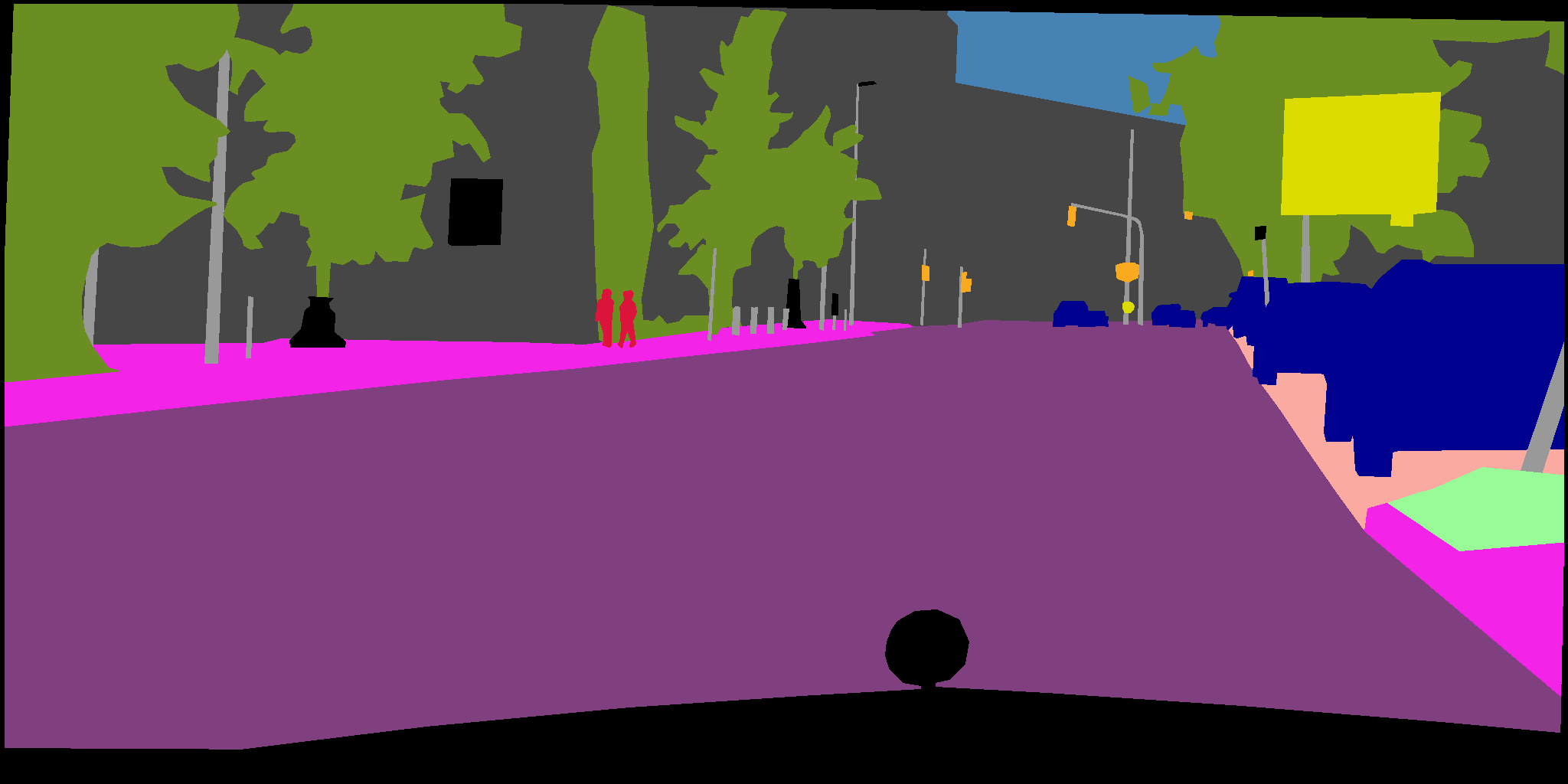} &
    \includegraphics[width=0.25\textwidth]{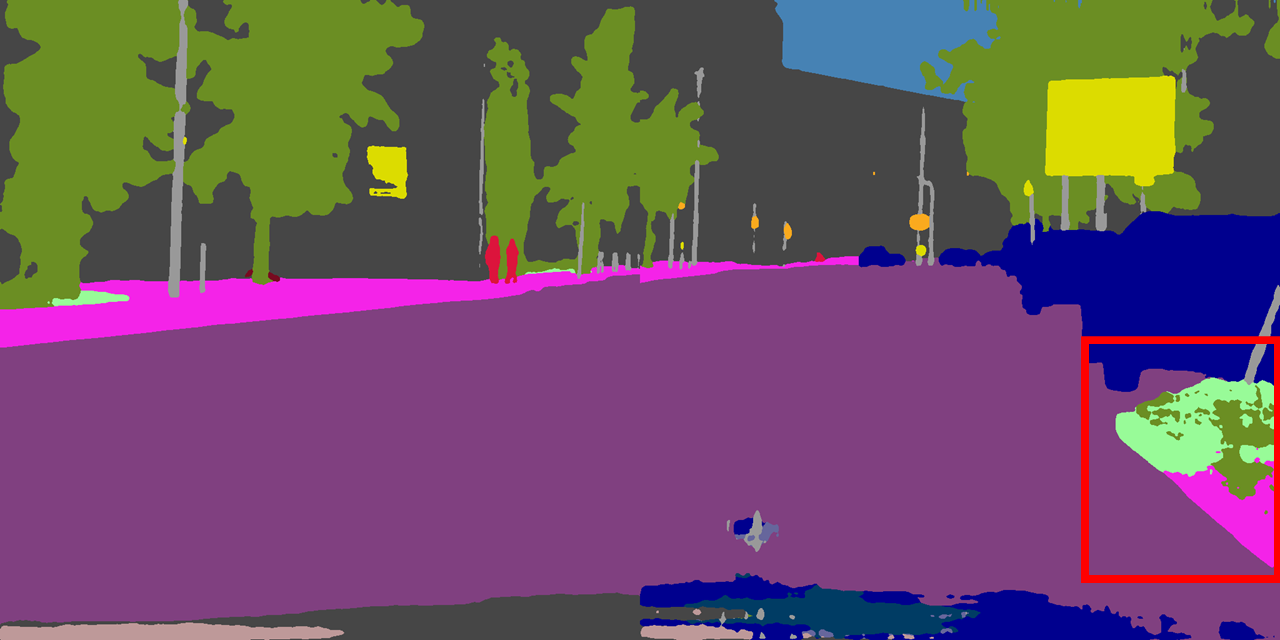} &
    \includegraphics[width=0.25\textwidth]{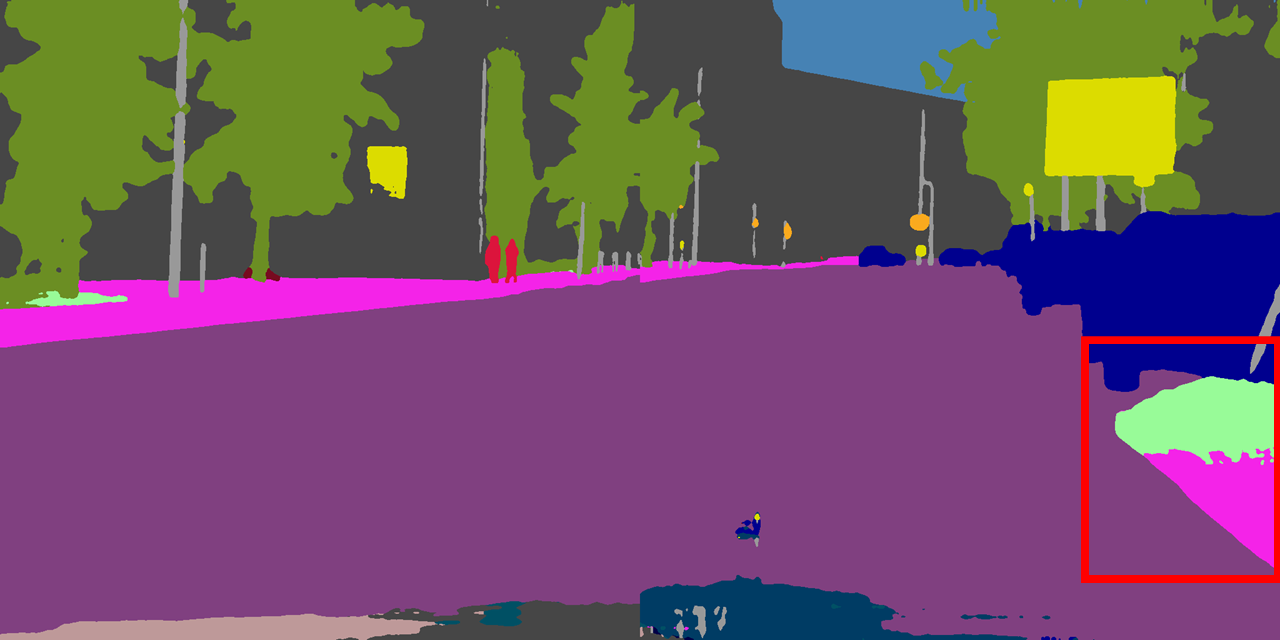} \\
    \includegraphics[width=0.25\textwidth]{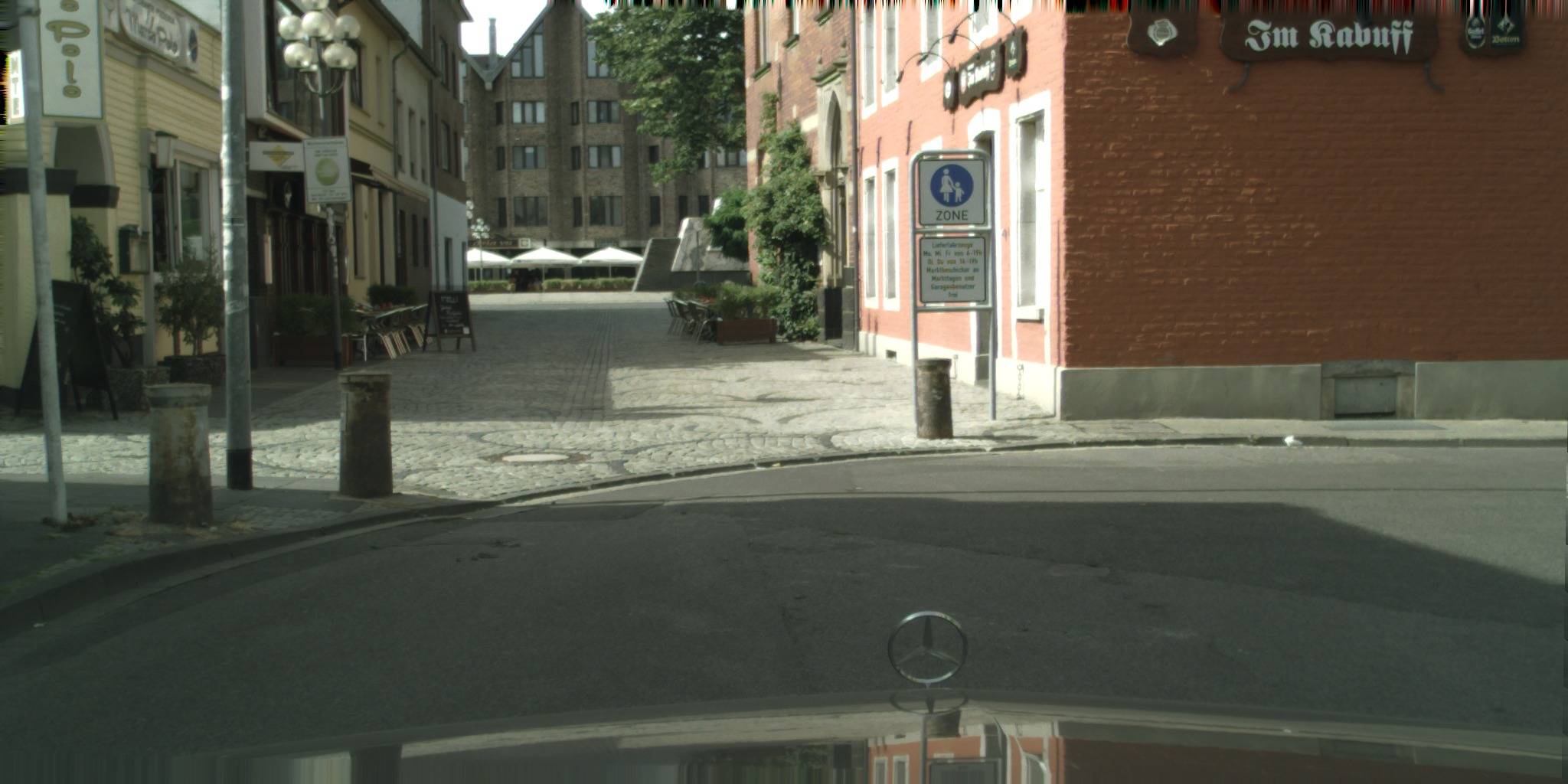} &
    \includegraphics[width=0.25\textwidth]{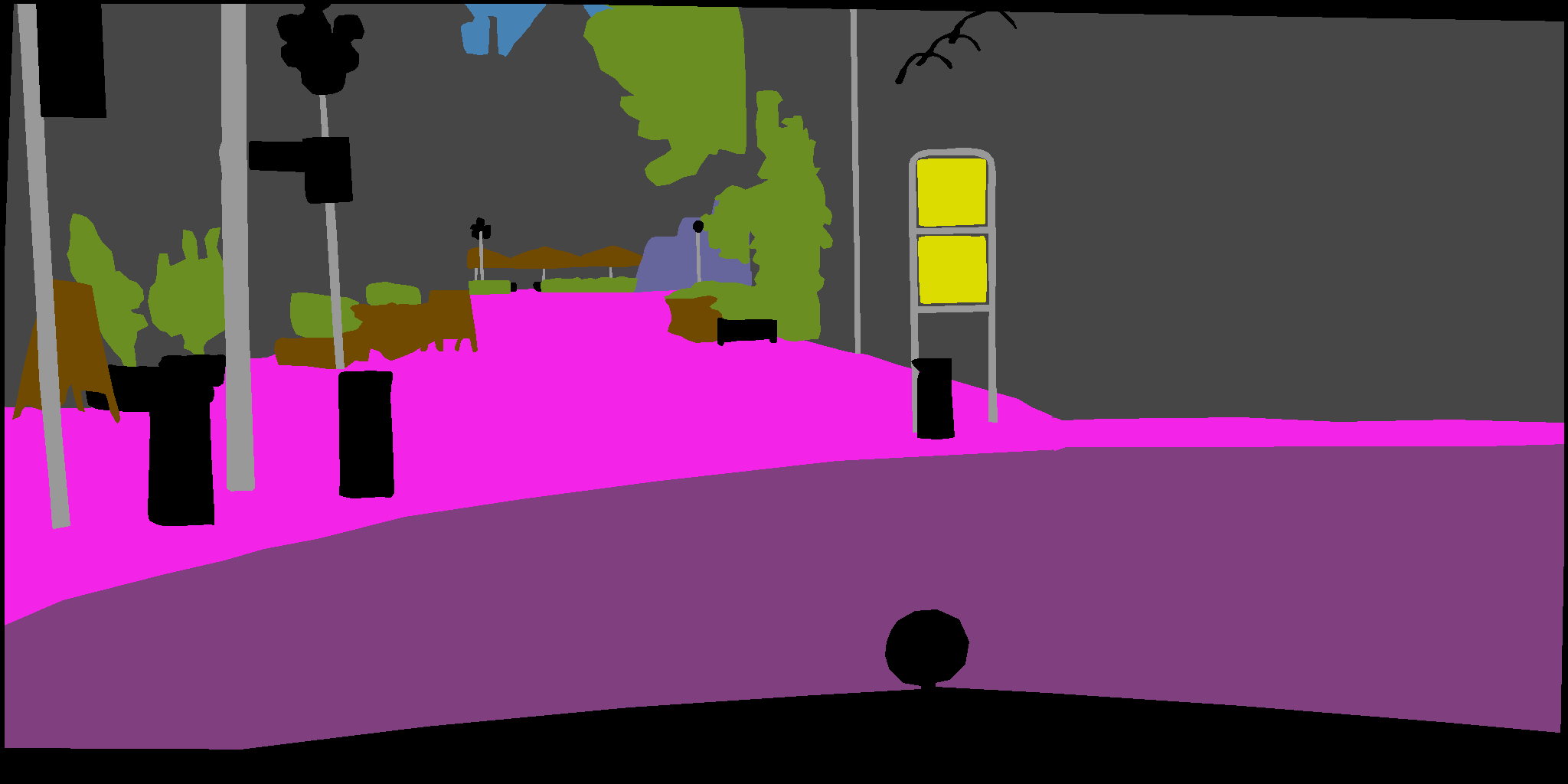} &
    \includegraphics[width=0.25\textwidth]{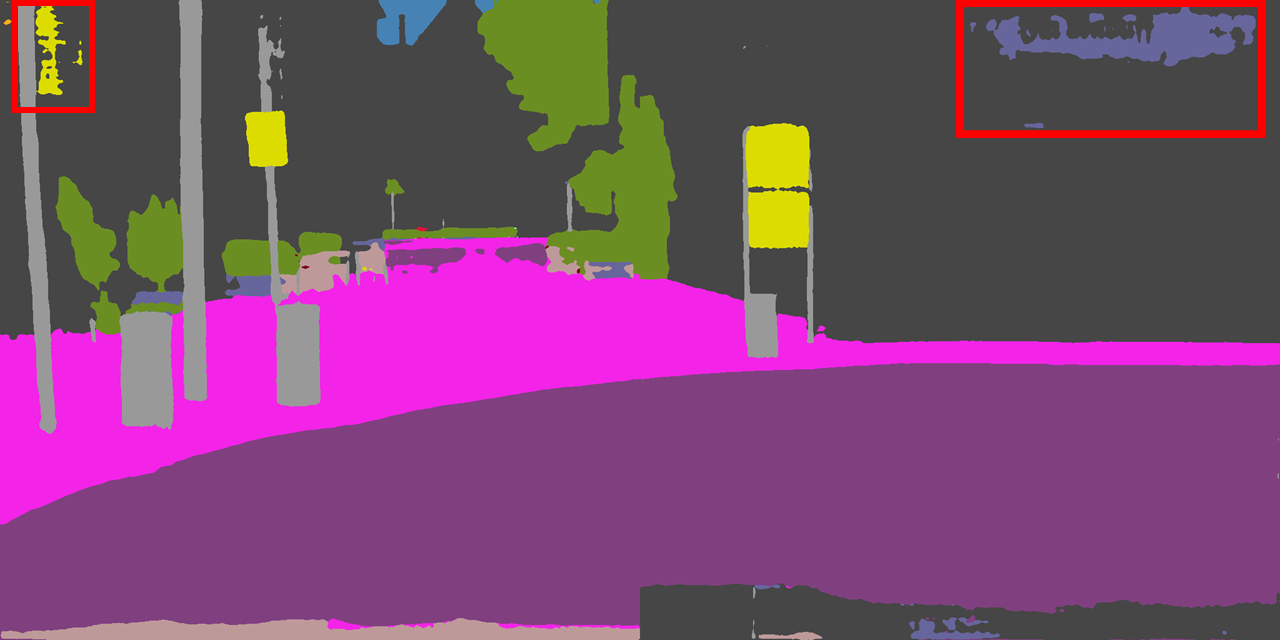} &
    \includegraphics[width=0.25\textwidth]{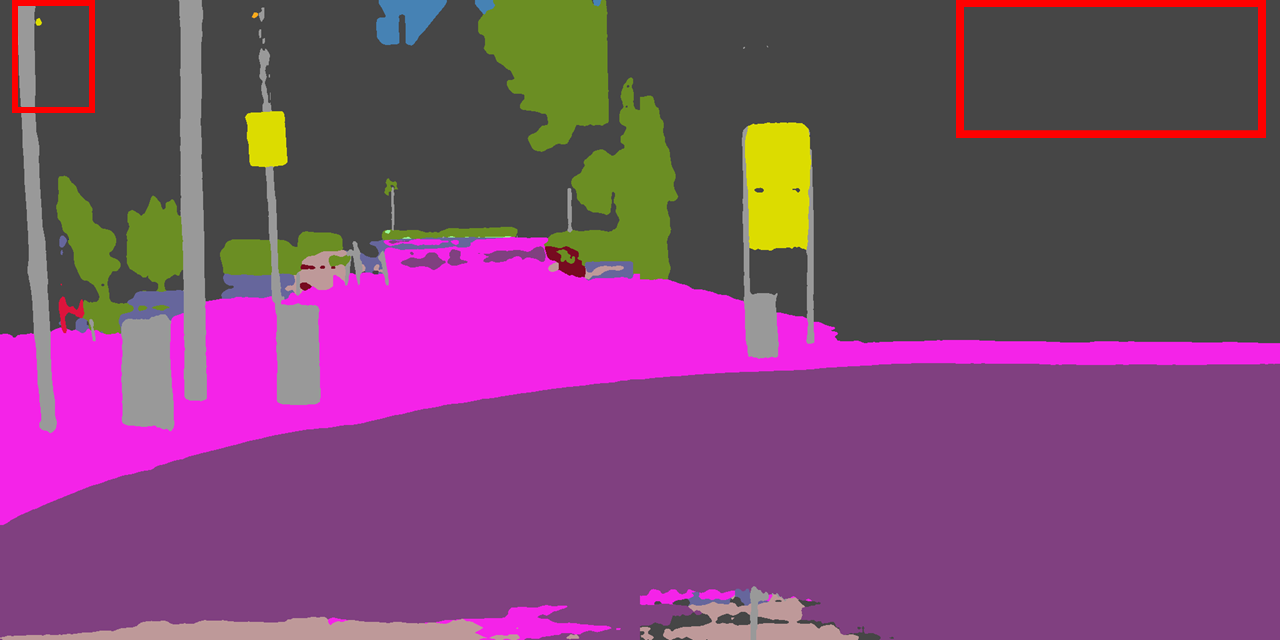} \\
    \includegraphics[width=0.25\textwidth]{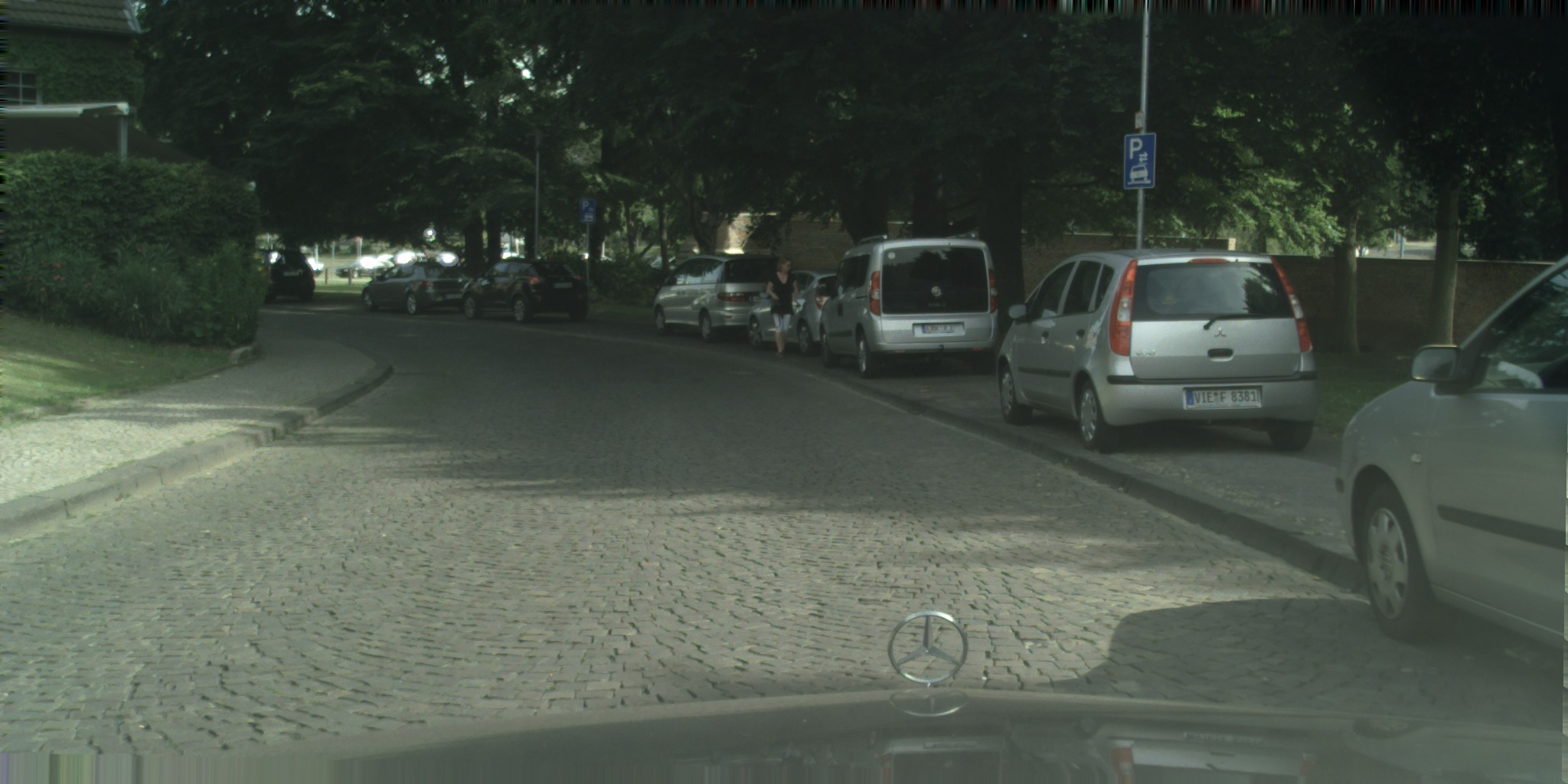} &
    \includegraphics[width=0.25\textwidth]{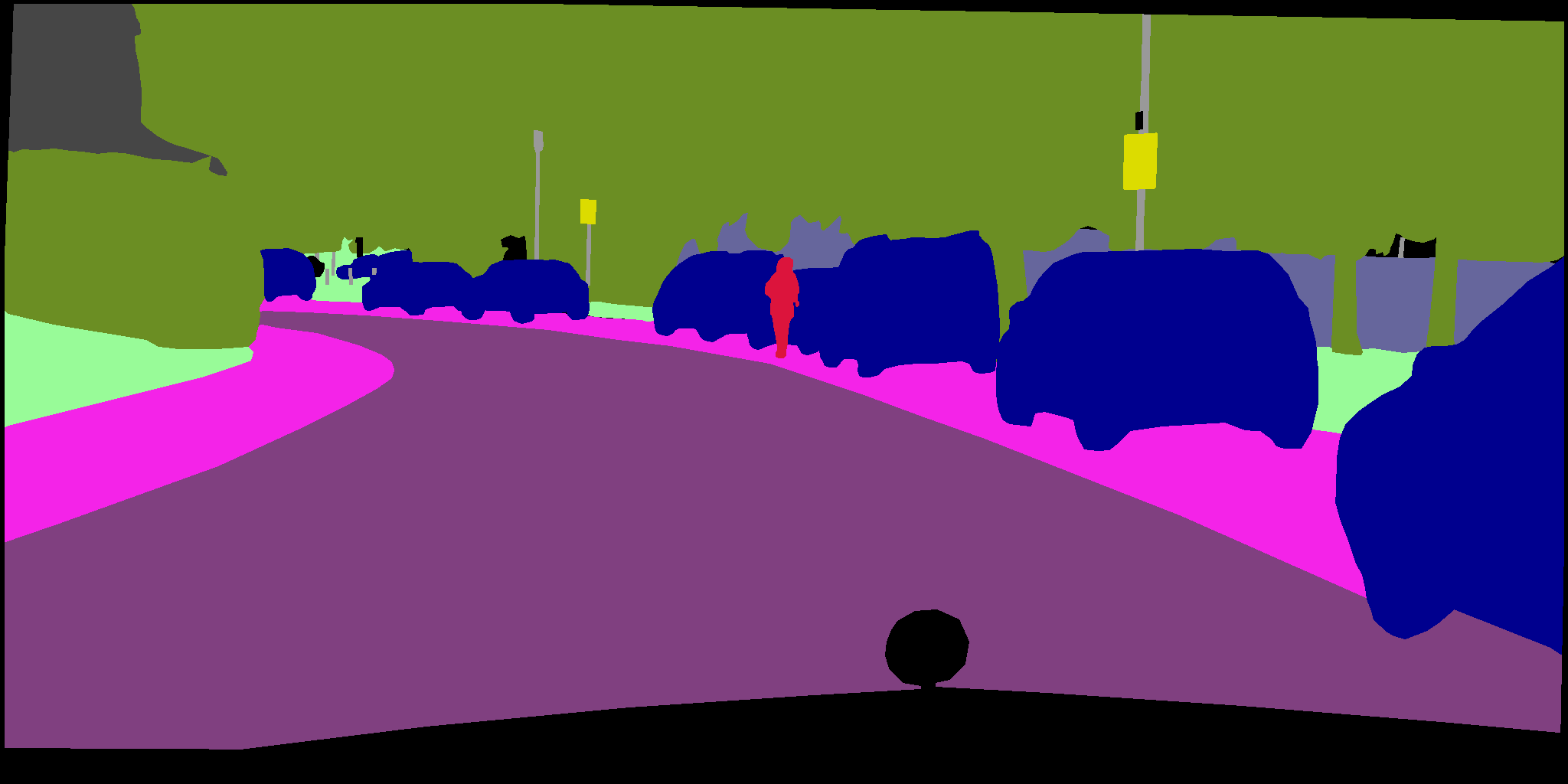} &
    \includegraphics[width=0.25\textwidth]{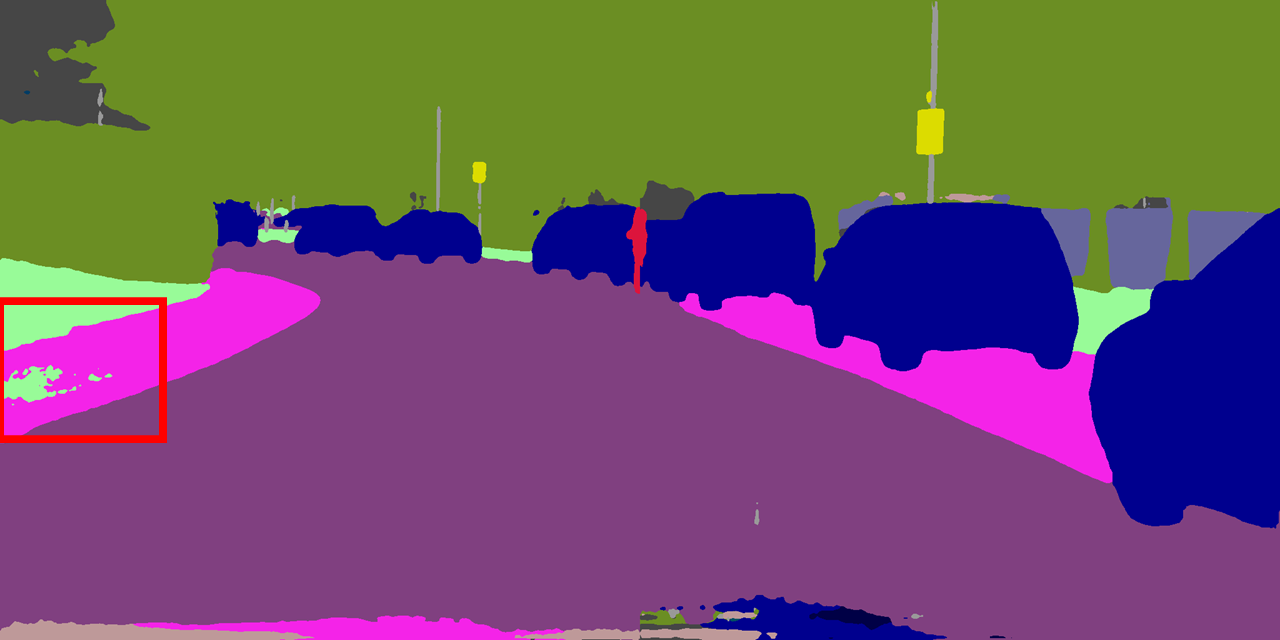} &
    \includegraphics[width=0.25\textwidth]{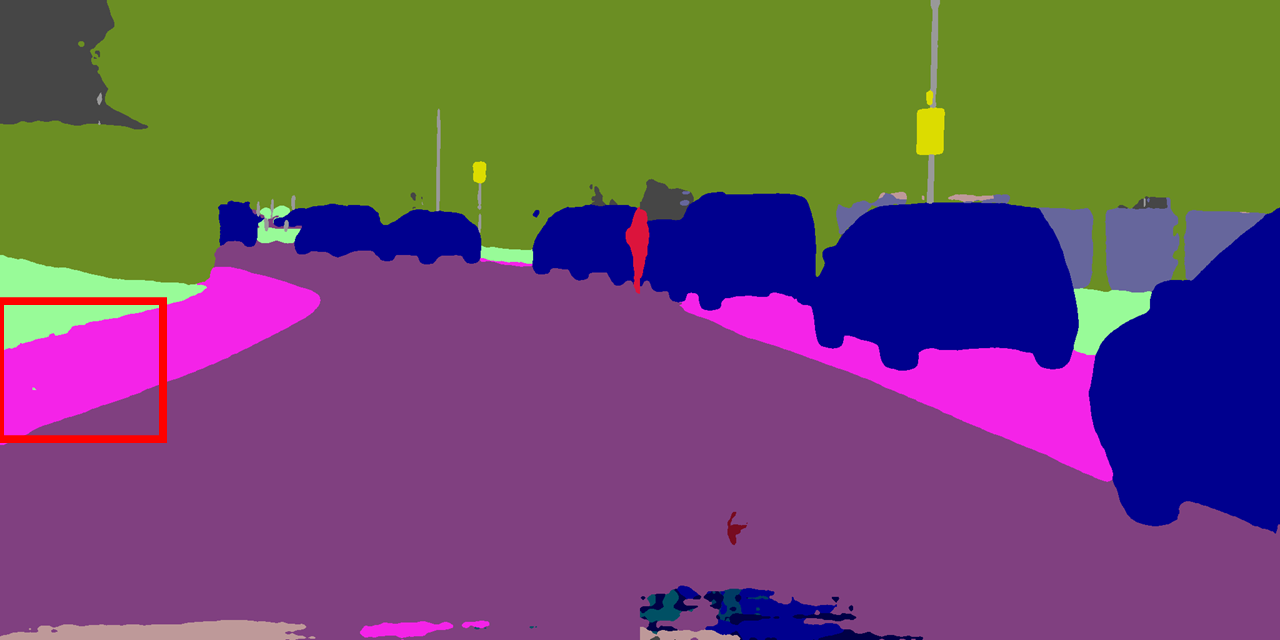} \\
    \includegraphics[width=0.25\textwidth]{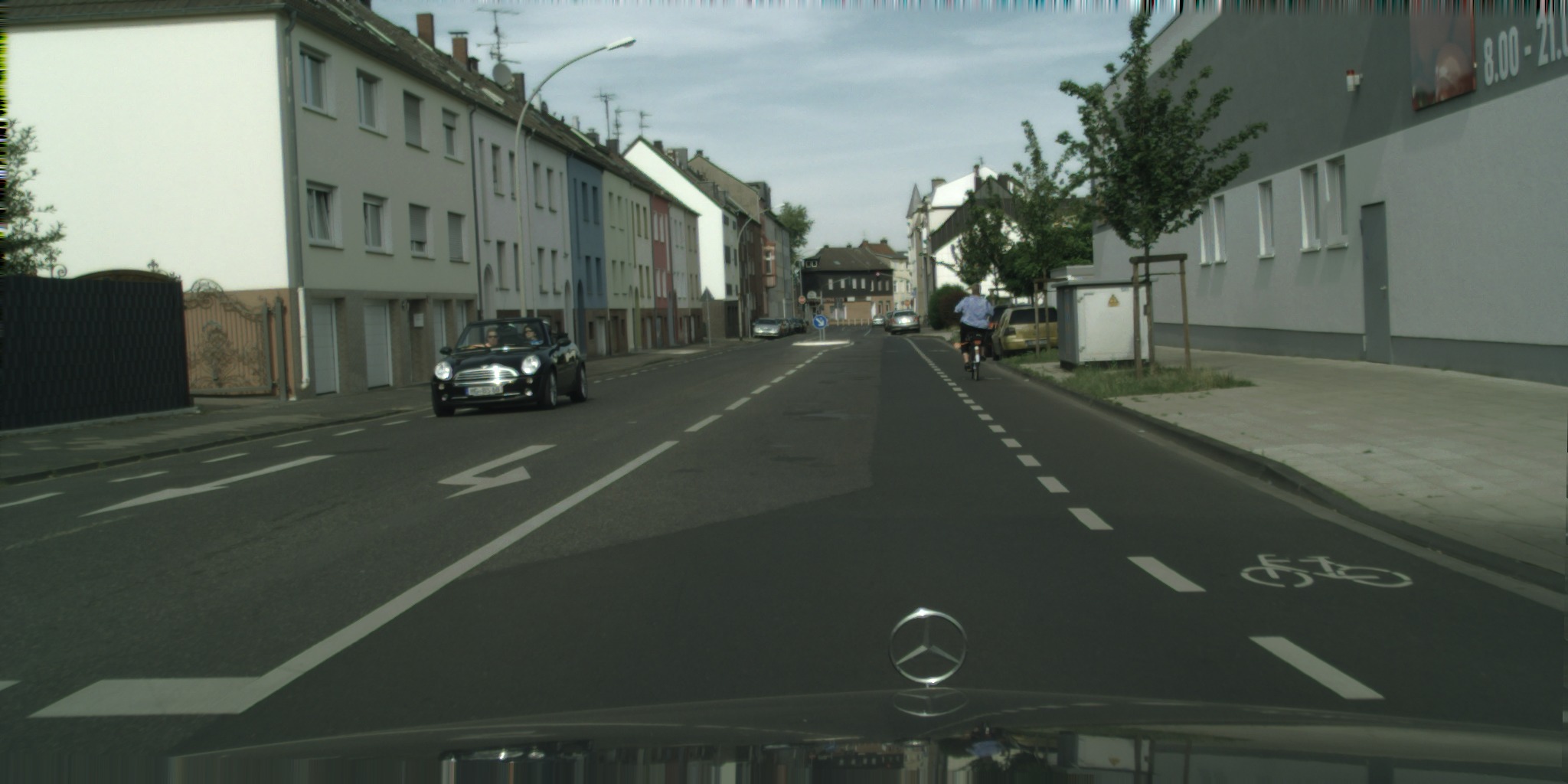} &
    \includegraphics[width=0.25\textwidth]{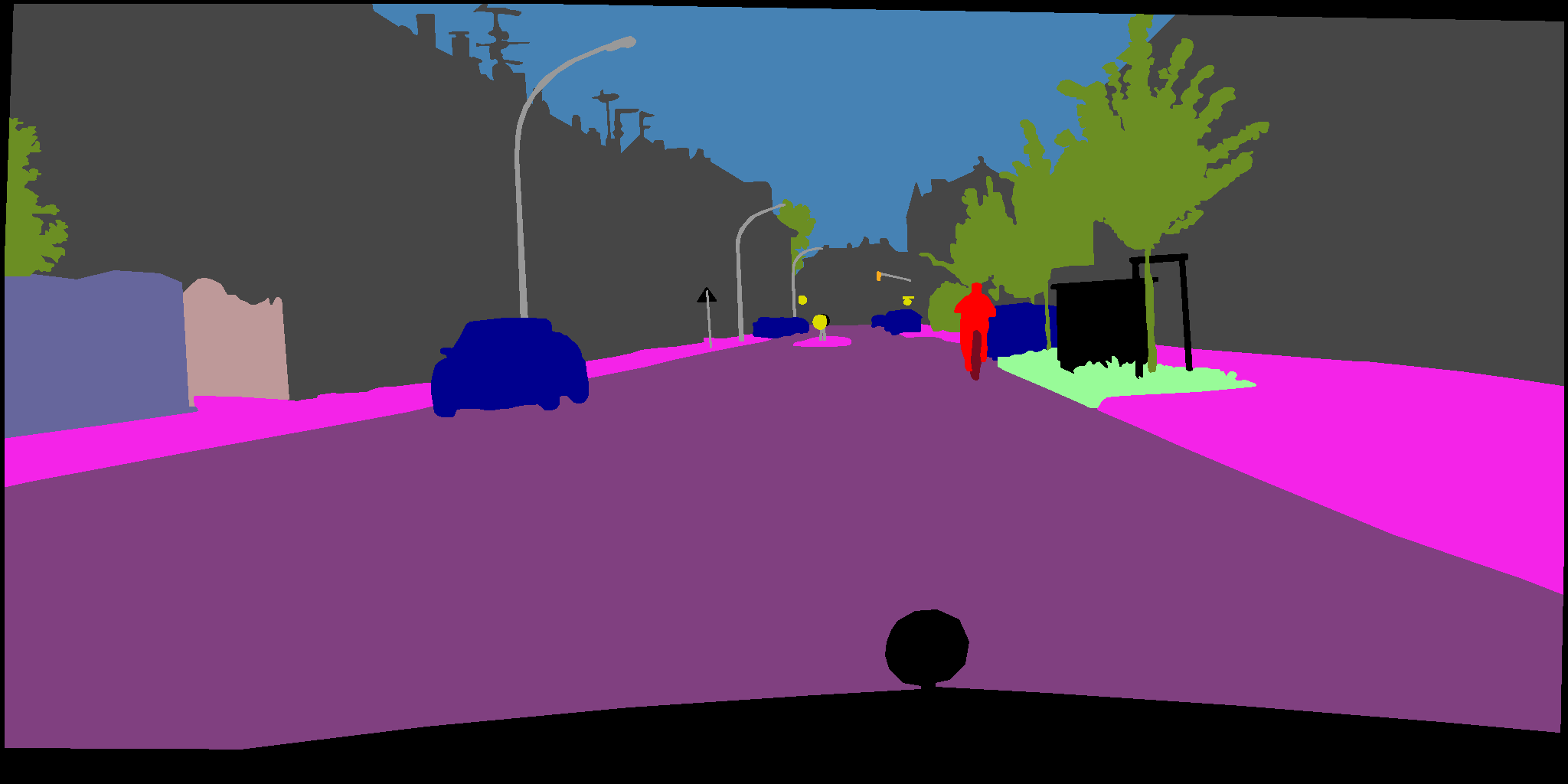} &
    \includegraphics[width=0.25\textwidth]{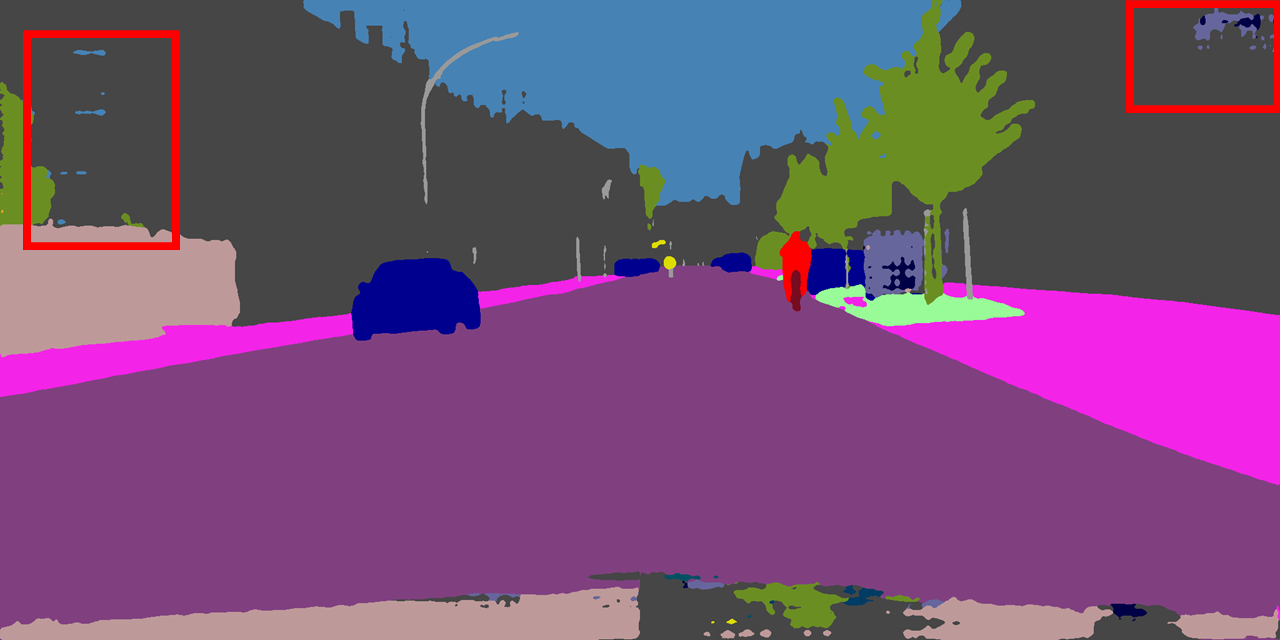} &
    \includegraphics[width=0.25\textwidth]{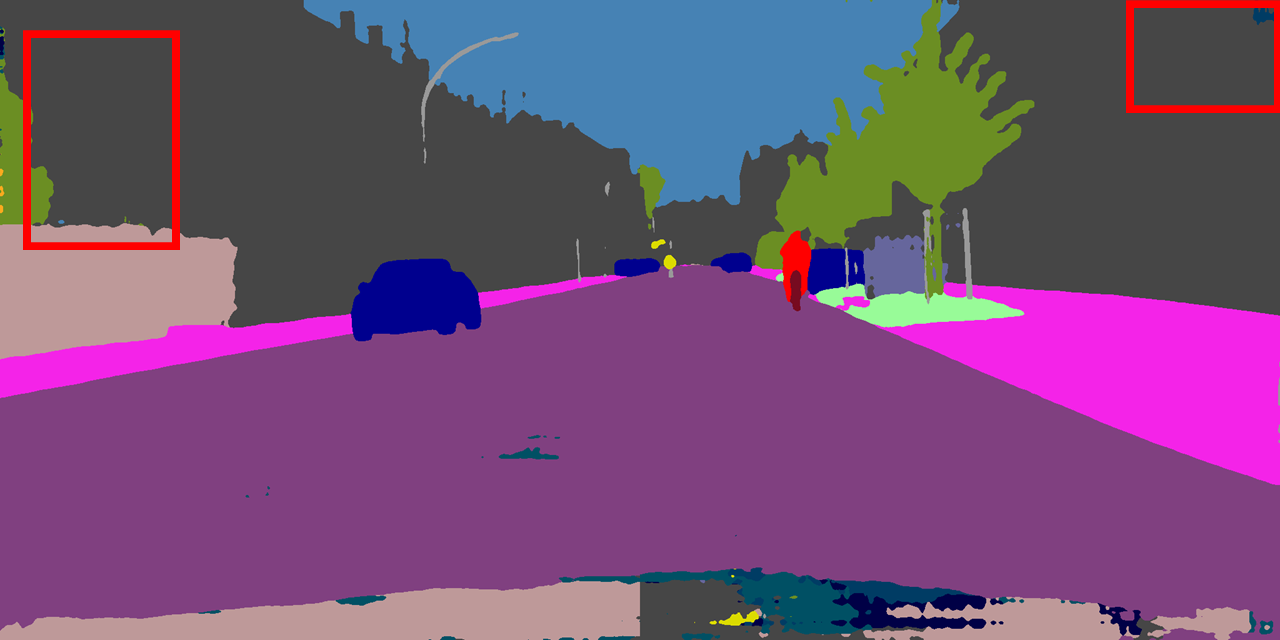} \\
    \includegraphics[width=0.25\textwidth]{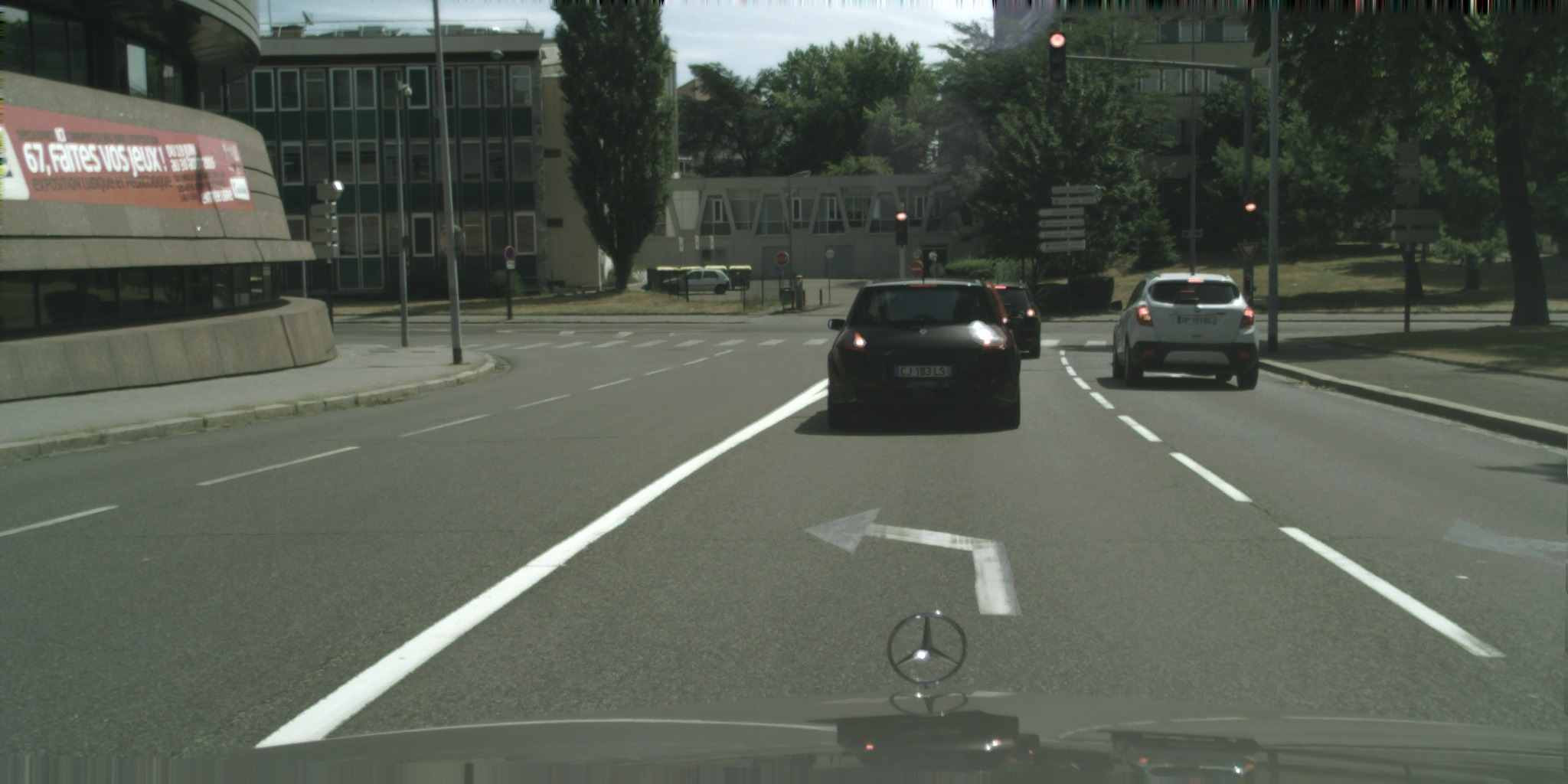} &
    \includegraphics[width=0.25\textwidth]{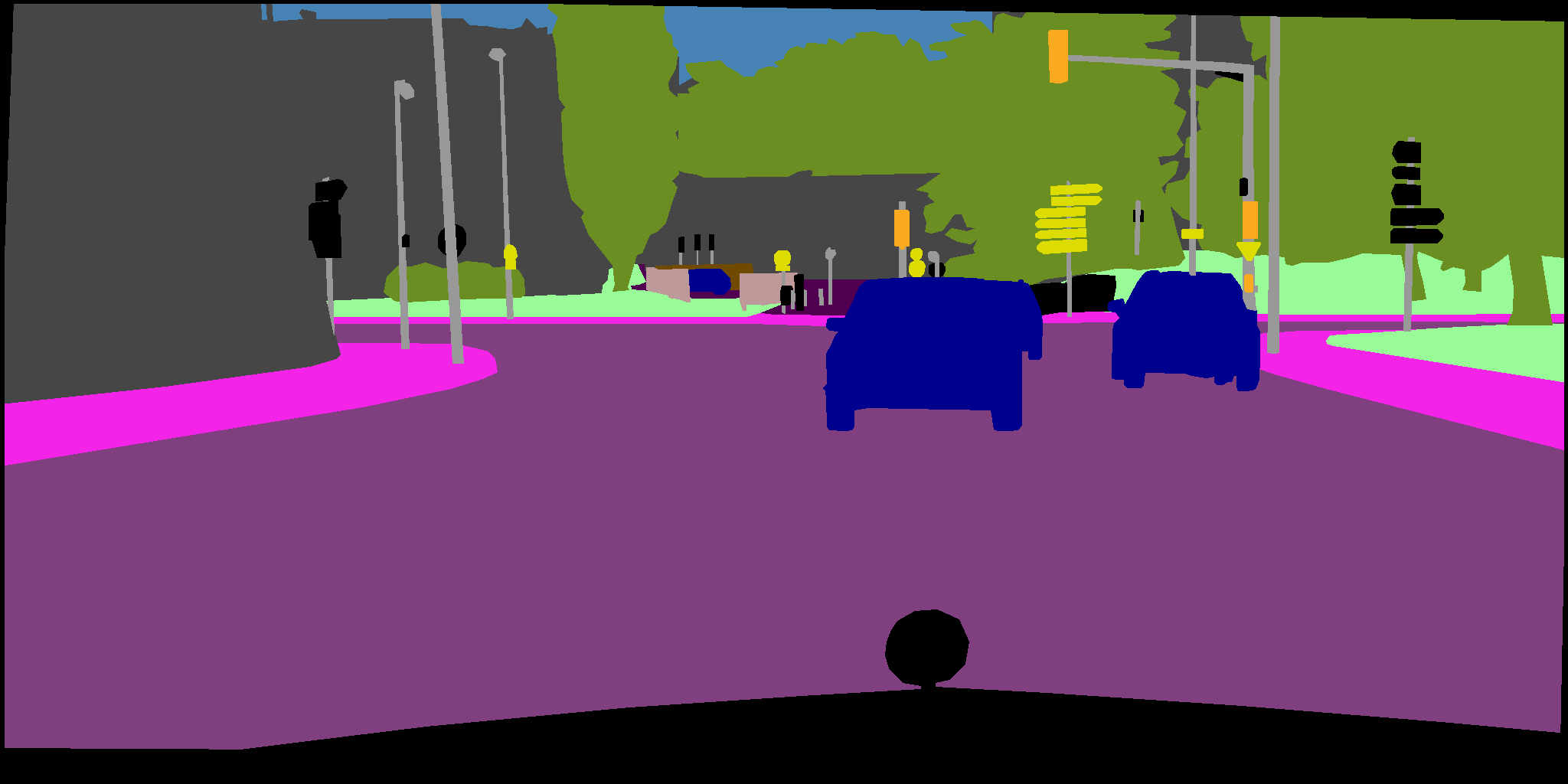} &
    \includegraphics[width=0.25\textwidth]{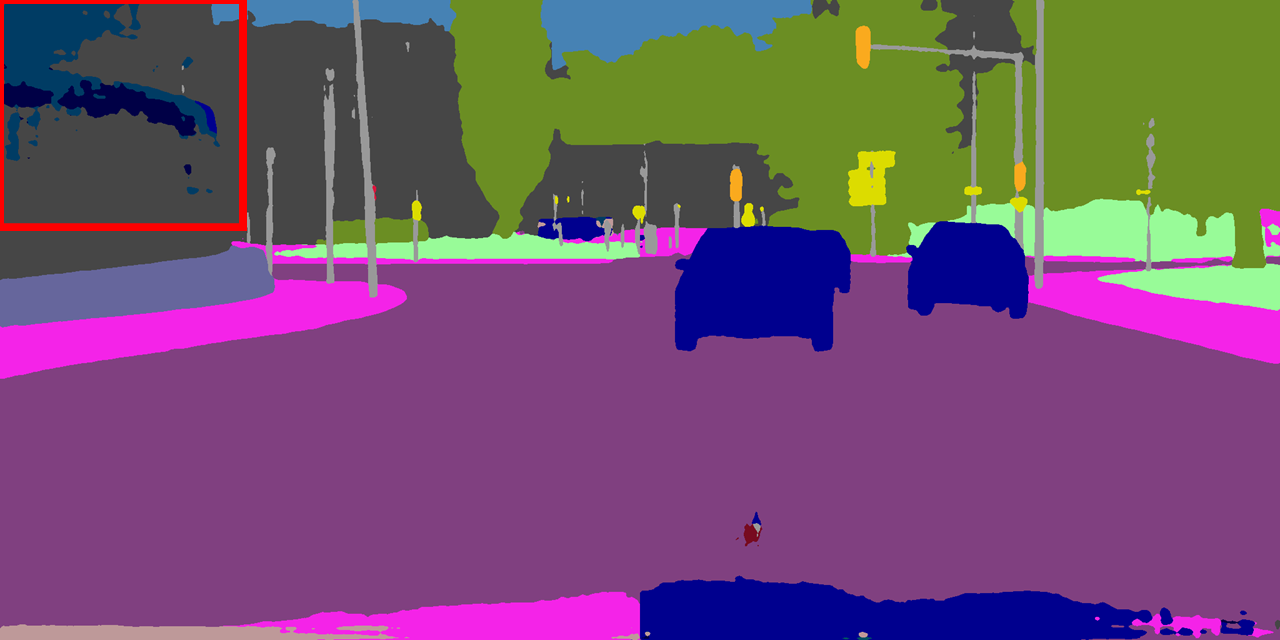} &
    \includegraphics[width=0.25\textwidth]{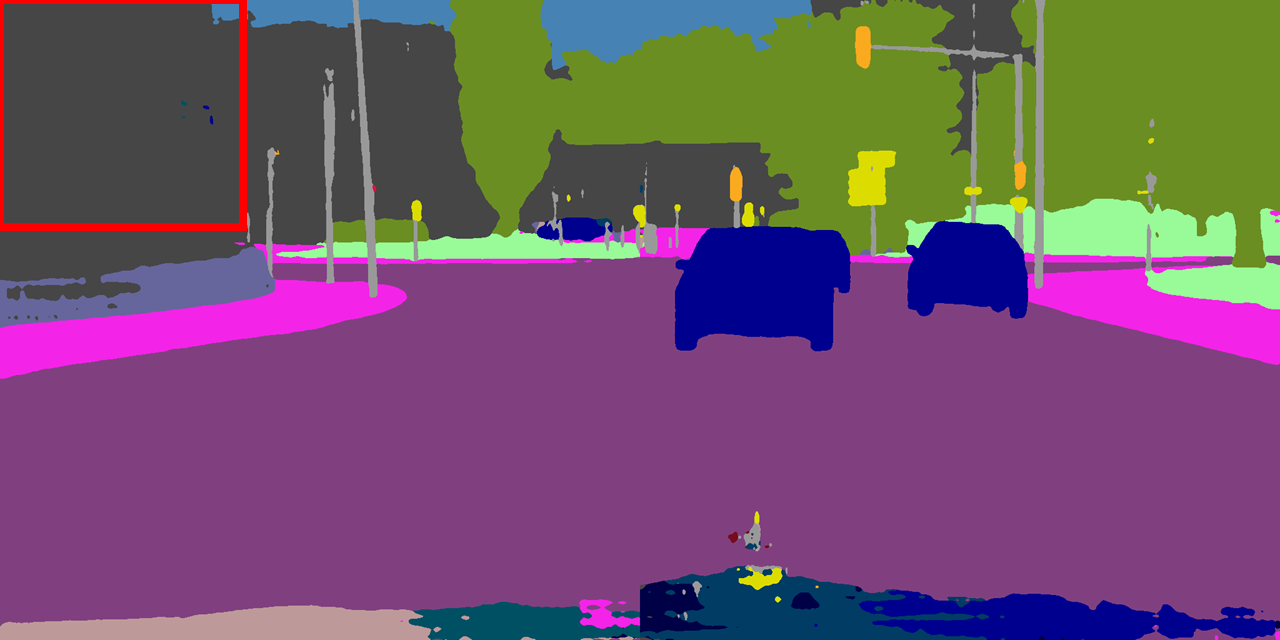} \\

\end{tabular}
}
\caption{Semantic segmentation using tile based evaluation without overlapping: Visual comparison on Cityscapes.  From left to right:  Image,  Ground Truth Segmentation, zero padding prediction, partial conv based padding prediction. We demonstrate that partial convolution based padding method can remove border  artifacts thus resulting in a better prediction.}
\label{fig:seg_result}
\vspace{-.5cm}
\end{figure*}

In Figure~\ref{fig:seg_result_0.3overlap}, we show segmentation comparisons between partial convolution based padding and zero padding for WideResnet38 for the tile based evaluation mode with $\frac{1}{3}$ overlapping. In Figure~\ref{fig:seg_result}, we show the segmentation comparisons for the tile based evaluation mode without overlapping. It can be seen that partial convolution based padding leads to better segmentation results on border regions.

\begin{figure*}
\centering
    \includegraphics[width=0.22\textwidth]{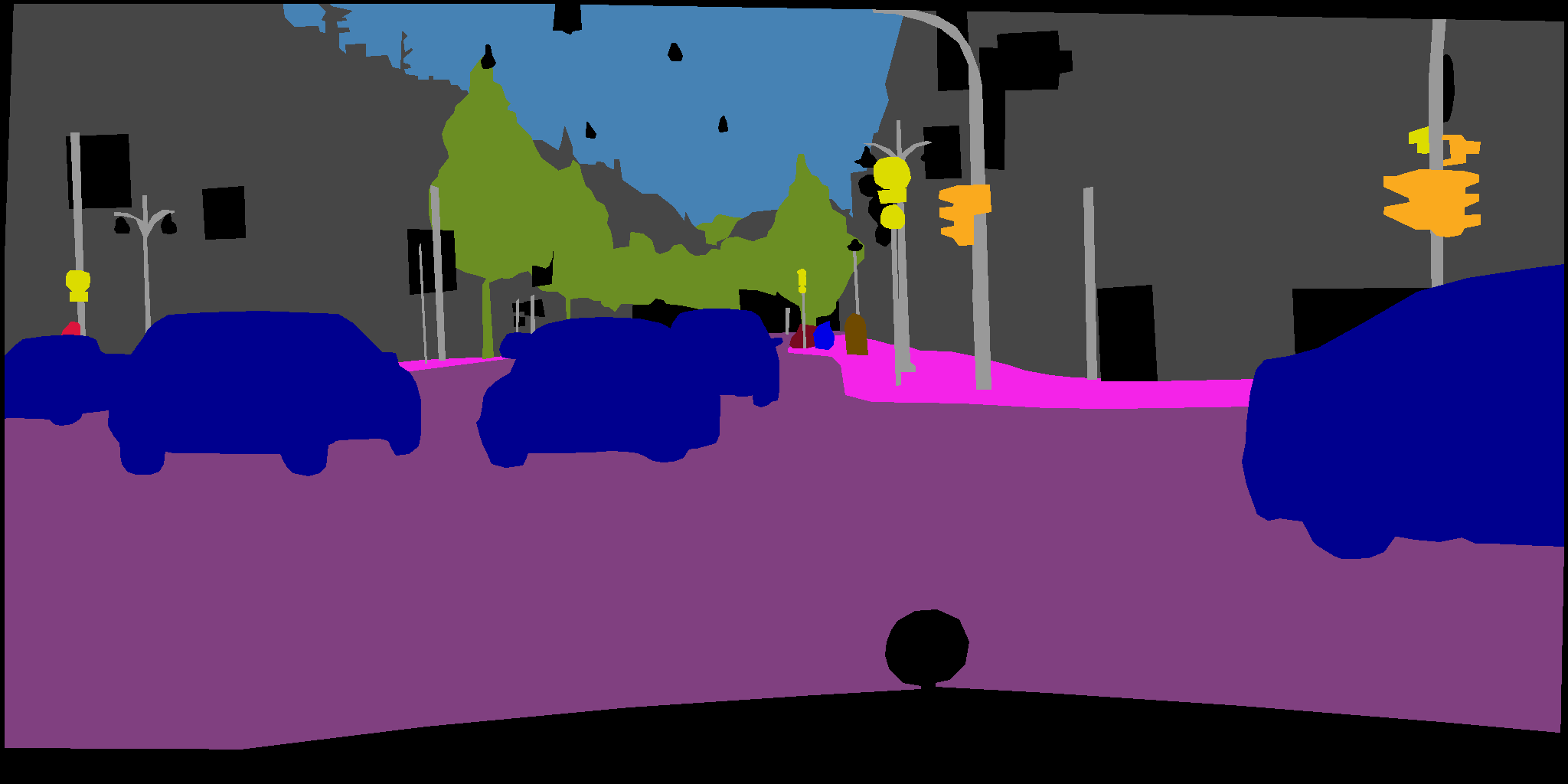}
    \includegraphics[width=0.22\textwidth]{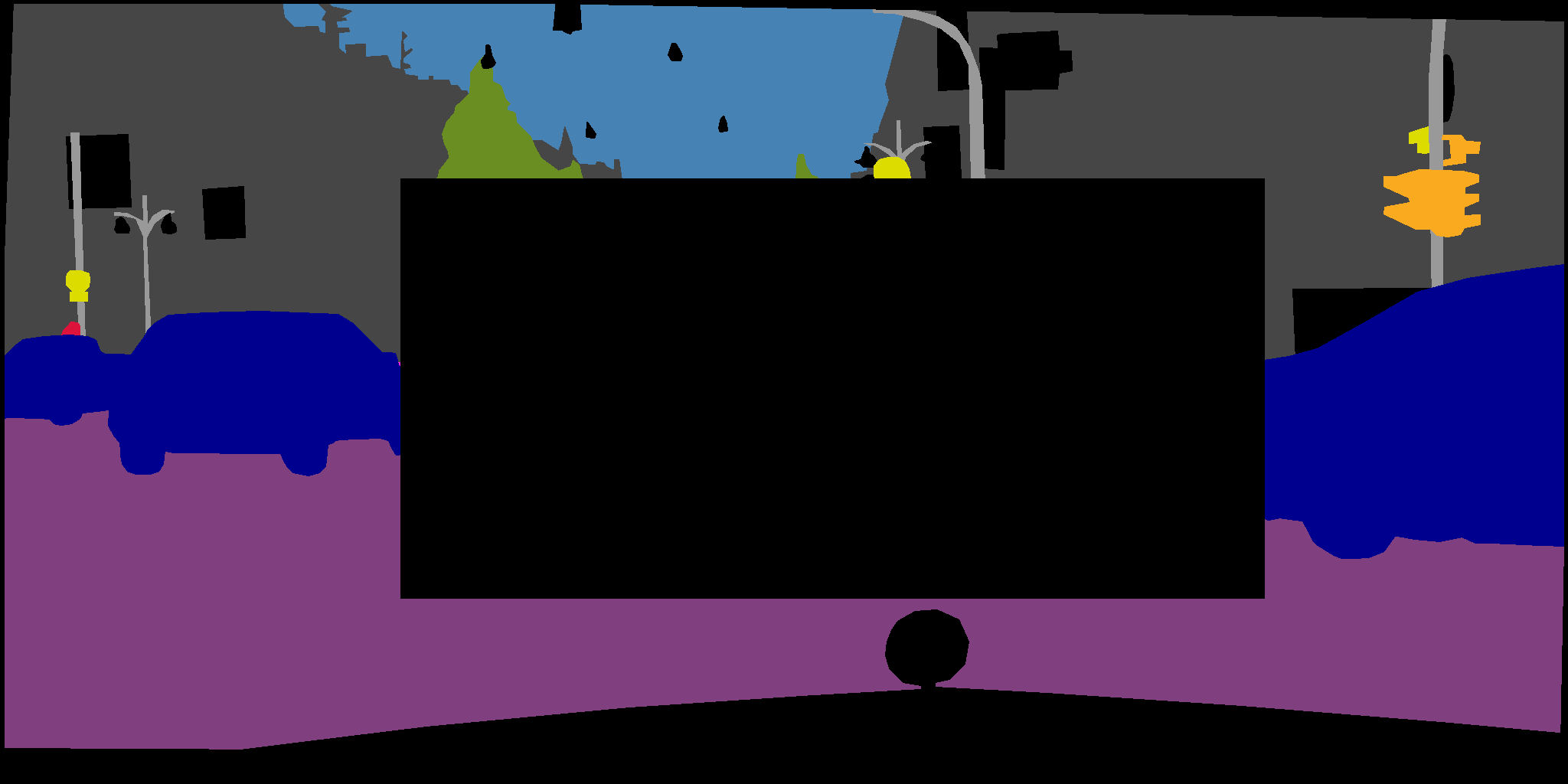}
    \includegraphics[width=0.22\textwidth]{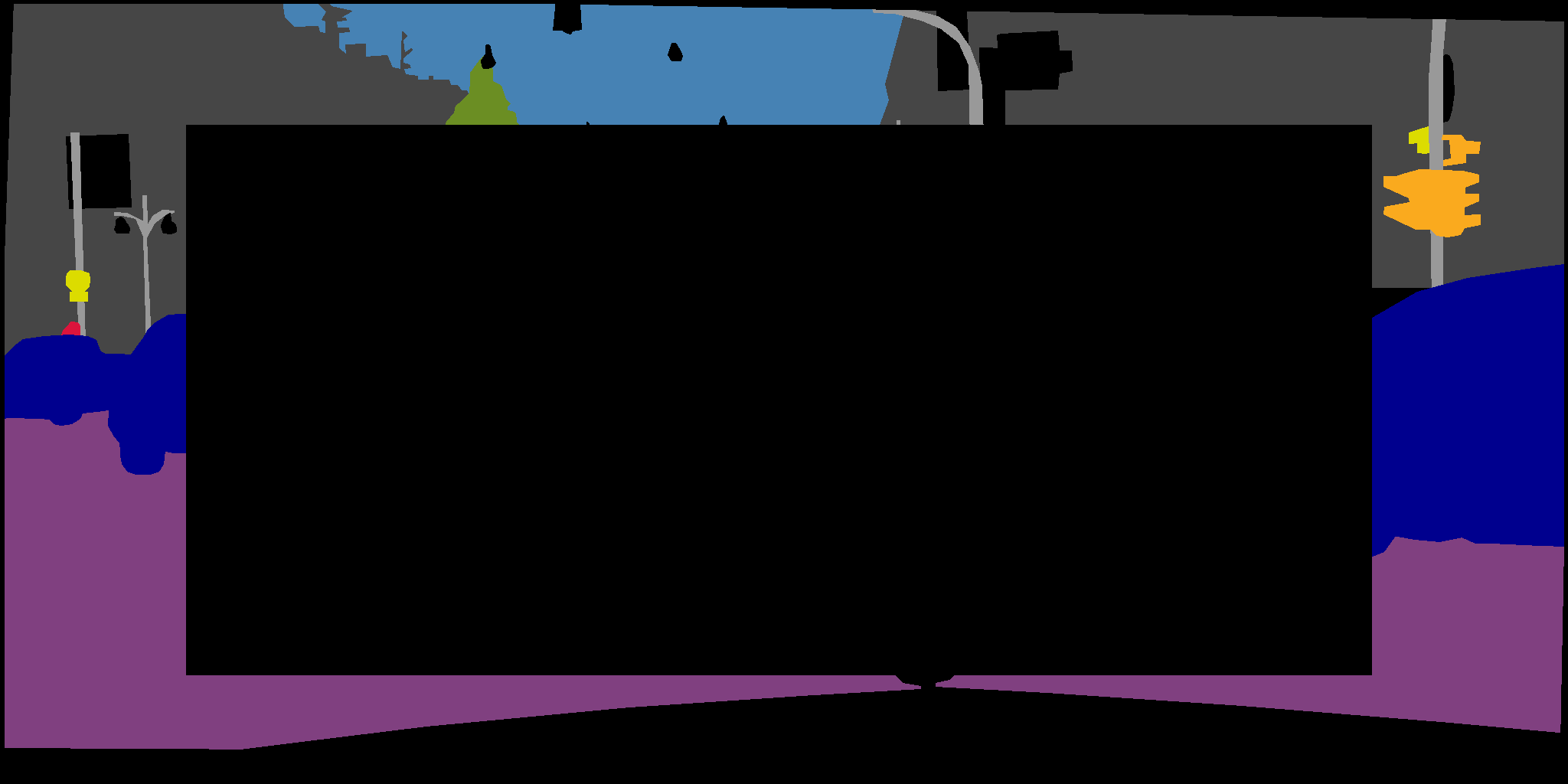}
    \includegraphics[width=0.22\textwidth]{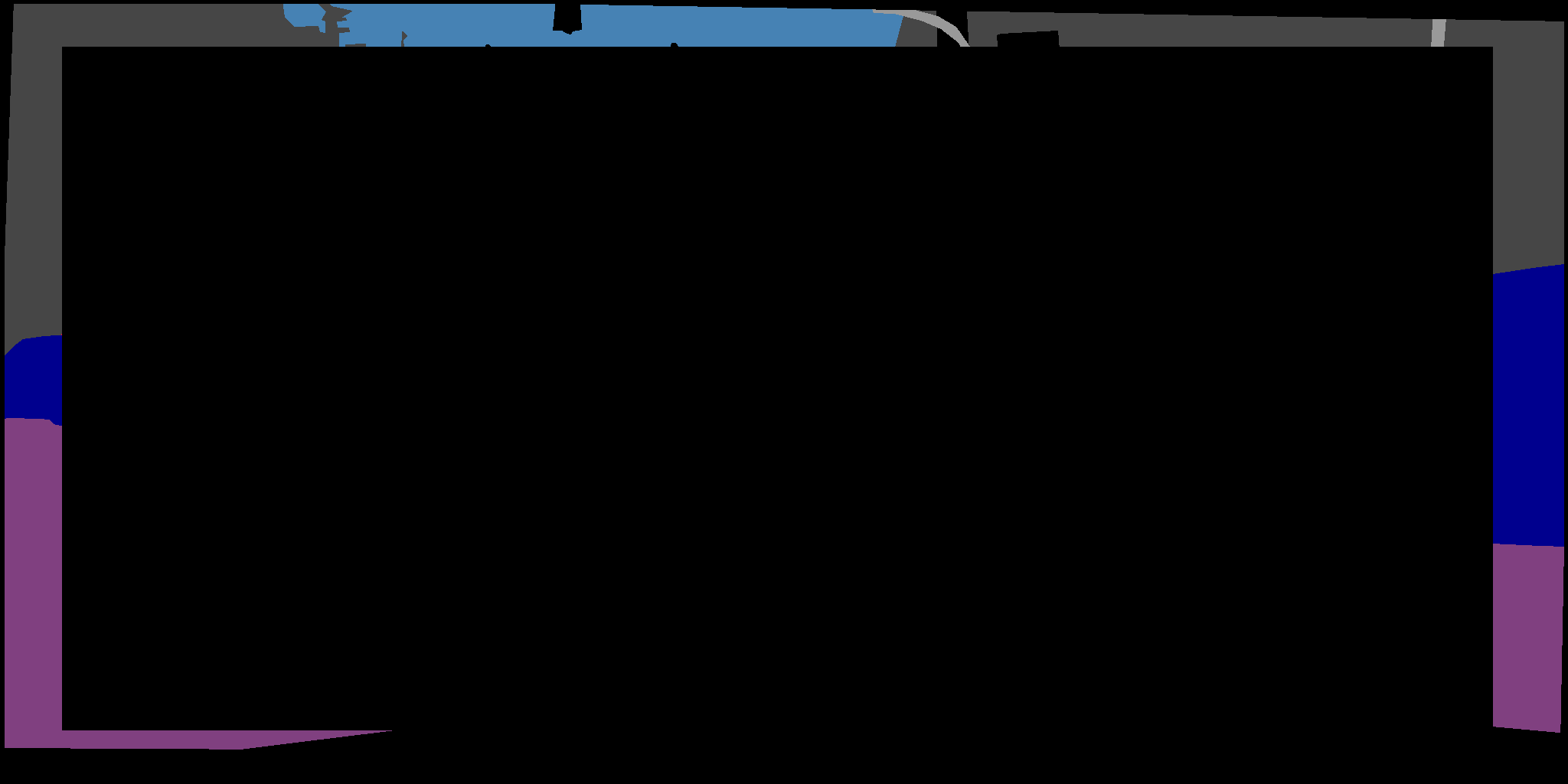} 
\caption{Samples with different proportions of leaving out center regions.}
\label{fig:centersample}
\vspace{-0.2cm}
\end{figure*}

\begin{table*}
    \centering
    \begin{tabular}{c|c|cccccc}
    \multicolumn{1}{c}{} & \multicolumn{1}{c}{} & \multicolumn{6}{|c}{center leave-out proportion} \\
    \hline
    padding & regular & 0 & $\frac{1}{3}\times\frac{1}{3}$ & $\frac{1}{2}\times\frac{1}{2}$ & $\frac{2}{3}\times\frac{2}{3}$ & $\frac{3}{4}\times\frac{3}{4}$ & $\frac{7}{8}\times\frac{7}{8}$ \\
    \hline
    WN38\_zero & 79.069 &  79.530 &  80.298 & 80.472 & 81.197 & 80.684 & 78.981 \\
    WN38\_partial & 79.082 & 79.907 & 80.683 & 81.051 & 82.031 & 81.600 & 80.230 \\   
    \hline
    diff & 0.012 & 0.377 & 0.385 & 0.579 & 0.834 & 0.916 & 1.249 \\
    \end{tabular}
    \caption{Evaluation of segmentation results by leaving out different proportions of the center region. Explanation can be found in Section~\ref{sec:border}.}
    \label{tab:leaveout}
    \vspace{-0.2cm}
\end{table*}

\subsubsection{Focused Evaluation on Border Regions}
\label{sec:border}

\textbf{Evaluation Excluding the Center Regions:} To better understand the advantage of partial convolution based padding in handling the border regions, we perform some additional evaluations which only evaluated the mIOUs on the border regions. Specifically, we set the target labels to ``don't care" for the center region of the image at varying proportions: $\frac{1}{3}\times\frac{1}{3}$, $\frac{1}{3}\times\frac{1}{3}$, $\frac{1}{2}\times\frac{1}{2}$, $\frac{2}{3}\times\frac{2}{3}$, $\frac{3}{4}\times\frac{3}{4}$, $\frac{7}{8}\times\frac{7}{8}$. Samples of different proportions of leaving out center regions can be found in Figure~\ref{fig:centersample}.
These evaluations use non-overlapping tiling. Similar to the Table 6 in the main paper, we select two WideResNet38-backbone models with different padding schemes but similar mIOU in the regular evaluation setting. Table~\ref{tab:leaveout} shows the corresponding evaluation results by leaving out different proportions of the center regions. It can be seen that as we leave out more proportions of the center region, the evaluation difference between zero padding and partial convolution based padding becomes larger. For example, if we only leave out $\frac{1}{3}\times\frac{1}{3}$ center regions, the partial convolution based padding only outperform the zero padding with $0.385\%$ mIOU. However, when we leave out $\frac{7}{8}\times\frac{7}{8}$ center regions, the difference becomes $1.249\%$. This further demonstrates that the partial-convolution based padding scheme significantly improves prediction accuracy near the image boundaries.

\section{Conclusion}
In this work, we present a new padding scheme called partial convolution based padding. We extensively evaluate the effectiveness of the partial convolution based padding compared to existing padding methods. We demonstrate that it outperforms the widely adopted zero padding through intensive experiments on image classification tasks and semantic segmentation tasks. We show that partial convolution based padding achieves better accuracy as well as faster convergence than the default zero padding on image classification. We also show that partial convolution based padding gets better segmentation results, especially at the image border, while the typical zero padding may reduce the model's certainty.

\noindent\textbf{Acknowledgement}
We would like to thank Jinwei Gu, Matthieu Le, Andrzej Sulecki, Marek Kolodziej and Hongfu Liu for helpful discussions. 

{\small
\bibliographystyle{ieee}
\bibliography{egbib}
}

\end{document}